\documentclass[sn-mathphys-num]{sn-jnl}

\usepackage{graphicx}%
\usepackage{multirow}%
\usepackage{amsmath,amssymb,amsfonts}%
\usepackage{amsthm}%
\usepackage{mathrsfs}%
\usepackage[title]{appendix}%
\usepackage{xcolor}%
\usepackage{textcomp}%
\usepackage{manyfoot}%
\usepackage{booktabs}%
\usepackage{algorithm}%
\usepackage{algorithmicx}%
\usepackage{algpseudocode}%
\usepackage{listings}%

\usepackage{adjustbox}
\usepackage{rotating} 
\usepackage{graphicx}

\theoremstyle{thmstyleone}%
\theoremstyle{thmstyletwo}%

\theoremstyle{thmstylethree}%

\begin{document}



\title[Article Title]{Beyond Diagnostic Performance: Revealing and Quantifying Ethical Risks in Pathology Foundation Models}


\author[1]{\fnm{Weiping} \sur{Lin}}\email{wplin@stu.xmu.edu.cn}
\equalcont{These authors contributed equally to this work.}

\author[1]{\fnm{Shen} \sur{Liu}}\email{liushen@stu.xmu.edu.cn}
\equalcont{These authors contributed equally to this work.}

\author[2]{\fnm{Runchen} \sur{Zhu}}\email{zhurunchen@stu.xmu.edu.cn}
\author[1]{\fnm{Yixuan} \sur{Lin}}\email{yixuanlin@stu.xmu.edu.cn}
\author[1]{\fnm{Baoshun} \sur{Wang}}\email{bswang@stu.xmu.edu.cn}
\author*[1]{\fnm{Liansheng} \sur{Wang}}\email{lswang@stu.xmu.edu.cn}

\affil*[1]{\orgdiv{Department of Computer Science at School of Informatics}, \orgname{Xiamen University}, \orgaddress{\city{Xiamen}, \postcode{361001}, \country{China}}}

\affil*[2]{\orgdiv{National Institute for Data Science in Health and Medicine}, \orgname{Xiamen University}, \orgaddress{\city{Xiamen}, \postcode{361001}, \country{China}}}




\abstract{
Pathology foundation models (PFMs), as large-scale pre-trained models tailored for computational pathology, have significantly advanced a wide range of applications. Their ability to leverage prior knowledge from massive datasets has streamlined the development of intelligent pathology models. 
However, we identify several critical and interrelated ethical risks that remain underexplored, yet must be addressed to enable the safe translation of PFMs from lab to clinic. These include the potential leakage of patient-sensitive attributes, disparities in model performance across demographic and institutional subgroups, and the reliance on diagnosis-irrelevant features that undermine clinical reliability. 
In this study, we pioneer the quantitative analysis for ethical risks in PFMs, including privacy leakage, clinical reliability, and group fairness. Specifically, we propose an evaluation framework that systematically measures key dimensions of ethical concern: the degree to which patient-sensitive attributes can be inferred from model representations, the extent of performance disparities across demographic and institutional subgroups, and the influence of diagnostically irrelevant features on model decisions. We further investigate the underlying causes of these ethical risks in PFMs and empirically validate our findings. Then we offer insights into potential directions for mitigating such risks, aiming to inform the development of more ethically robust PFMs. 
This work provides the first quantitative and systematic evaluation of ethical risks in PFMs. Our findings highlight the urgent need for ethical safeguards in PFMs and offer actionable insights for building more trustworthy and clinically robust PFMs. To facilitate future research and deployment, we will release the assessment framework as an online toolkit to support the development, auditing, and deployment of ethically robust PFMs.
}

\keywords{Pathology Foundation Models, Quantifying Ethical Risks, Lab-to-clinic Evaluation, Privacy Leakage, Clinical Reliability, Group Fairness}



\maketitle

\section{Introduction}\label{introduction}
Pathological examination offers definitive insights into the cellular and tissue-level manifestations of complex diseases such as cancers and chronic inflammatory conditions, serving as the gold standard for diagnosis and clinical decision-making \cite{song2023artificial,cui2021artificial}. With the advent of digital pathology, these examinations are typically transformed into high-resolution whole slide images (WSIs), often exceeding 10 billion pixels at $40\times$ magnification \cite{quellec2017multiple, gadermayr2024multiple}. Due to their immense size, few devices can process them directly, even when multiple instance learning models are employed. To address this, pretrained patch-level encoders are commonly employed to extract semantic features from image regions of interest. When such an encoder yields representations that generalize well across diverse diagnostic tasks, it is referred to as a Pathology Foundation Model (PFM). These models are typically trained on large-scale pathology image datasets and are capable of producing discriminative features that can be adapted to a variety of downstream clinical applications \cite{chanda2024new}. The advent of PFMs has significantly streamlined the development of deep learning solutions in computational pathology (CPath) by efficiently leveraging prior knowledge and improving diagnostic performance across heterogeneous scenarios.

In recent years, PFMs have rapidly advanced and can be broadly categorized into three main types. The first category includes models pretrained on natural image datasets, such as ResNet50 and DenseNet \cite{he2016deep, huang2017densely}, which are commonly repurposed as feature extractors in pathology tasks. The second category consists of models pretrained on pathology image patches using self-supervised learning. Notable examples include CTransPath \cite{wang2022transformer}, UNI \cite{chen2024towards}, Prov-GigaPath \cite{xu2024whole}, and Virchow \cite{vorontsov2024foundation}, which learn discriminative representations from large-scale pathology images. The third category comprises multi-modal PFMs, including PLIP \cite{huang2023visual}, CONCH \cite{lu2024visual}, and PathChat \cite{lu2024multimodal}, where the visual branch serves as a highly effective feature extractor. These PFMs have delivered impressive results across a wide range of downstream applications, including tumor subtype classification, grading, mutation prediction, biomarker prediction, and prognosis estimation.

Despite the remarkable progress and broad applicability of PFMs in CPath, they still present critical ethical risks that may hinder their safe and effective translation from laboratory settings to real-world clinical practice. If unaddressed, these risks could undermine the reliability, fairness, and trustworthiness of PFM-powered diagnostic systems, thereby limiting their clinical utility. In this study, we focus on three key dimensions of ethical concern. The first ethical concern is privacy risk. PFMs may inadvertently encode sensitive patient attributes (e.g., age, gender, race) and the medical institution into their feature representations. This raises serious concerns about patient privacy, especially when such information can be recovered from intermediate model outputs. In clinical settings where confidentiality is paramount, the leakage of demographic or institutional identifiers can lead to unintended disclosures and undermine trust in AI-driven diagnostics. The second ethical concern is clinical reliability. The ultimate goal of PFM-based systems is to assist physicians in making accurate decisions across diverse clinical environments. However, these models may rely on features that are not causally linked to disease pathology. Such diagnostic-irrelevant information may exhibit spurious correlations with labels during training, but fail to generalize across hospitals, devices, or patient populations. This compromises the robustness and reproducibility of the system in real-world deployment. The third ethical concern is fairness. Clinical populations are inherently heterogeneous, encompassing diverse demographic groups and healthcare institutions. Ethical AI systems must ensure equitable performance across these subgroups. Significant disparities in diagnostic accuracy across gender, race, or institutional subpopulations not only introduce clinical bias, but also risk exacerbating existing healthcare inequalities. 


Although certain ethical concerns have been widely discussed in previous studies, for example, Hanna et al. reviewed ethical and bias considerations in artificial intelligence and machine learning \cite{hanna2025ethical}, and Du et al. summarized potential ethical challenges of PFMs such as algorithmic bias, data privacy, and misuse \cite{du2025ethics}, most of them remain theoretical and lack quantitative evaluation methods. In addition, some studies have assessed issues beyond diagnostic performance. For instance, Seyyed et al. \cite{seyyed2021underdiagnosis}, Vaidya et al. \cite{vaidya2024demographic}, and Yang et al. \cite{yang2025demographic} discussed demographic bias in medical AI models, and Gustafsson et al. found that although foundation models outperformed baseline models in prostate cancer grading tasks, they still suffered significant performance degradation under common distribution shifts \cite{gustafsson2024evaluating}. On the other hand, several recent efforts have focused on benchmarking PFMs. During the development of PFMs, researchers have validated them across a broad range of downstream tasks. Campanella et al. constructed a benchmark dataset encompassing cancer diagnosis and various clinically relevant biomarkers, along with an automated evaluation pipeline \cite{campanella2025clinical}. Ma et al. proposed PathBench, a comprehensive benchmark for evaluating PFMs in precision oncology \cite{ma2025pathbench}. However, these efforts primarily focus on diagnostic accuracy across tasks and do not address the quantification of ethical risks. In summary, ethical risk assessment for PFMs still faces several limitations: (1) discussions around ethical risks are largely fragmented and theoretical, lacking practical and quantifiable methodologies; and (2) there is a shortage of clinically relevant and comprehensive quantitative assessment frameworks from an ethical standpoint.


\begin{figure}[htp]
\centering
\includegraphics[width=0.90\textwidth]{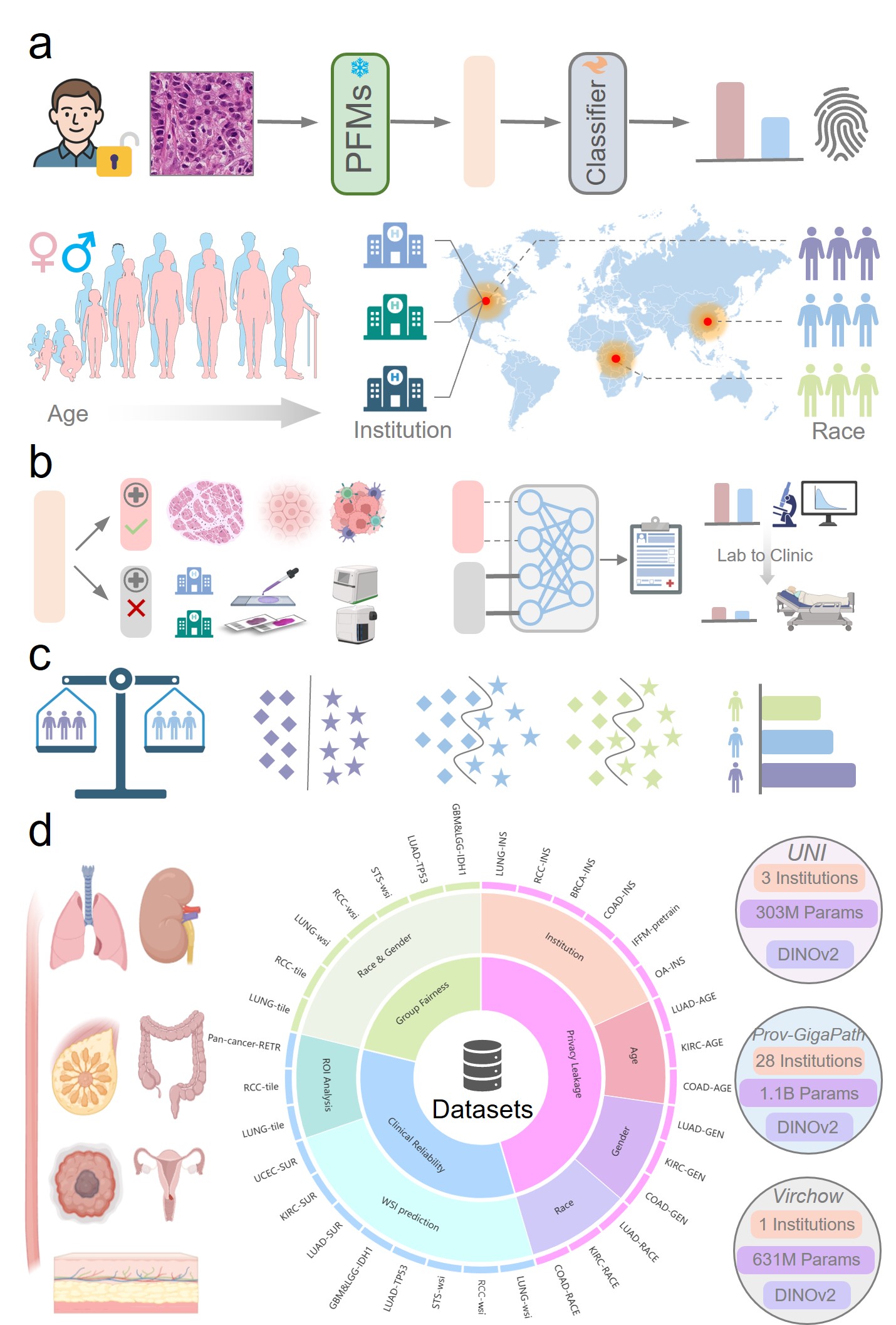}
\caption{Overview of the ethical risk evaluation framework. Pathology images are first processed by PFMs to obtain feature embeddings. \textbf{a}. The potential leakage of patient-sensitive attributes is assessed by evaluating whether information such as gender, age, race, and hospital origin can be inferred from the embeddings. \textbf{b}. The influence of diagnostically irrelevant features on model decisions is measured by quantifying the performance degradation of downstream models under distribution shifts. \textbf{c}. Group fairness is evaluated by analyzing the separability of embeddings and the final diagnostic accuracy across demographic and institutional subgroups under the same task. \textbf{d.} The datasets and pathology foundation models. }
\label{fig:fig1}
\end{figure}

To address these critical yet underexplored challenges, we pioneer the quantitative analysis of ethical risks in PFMs. In this study, we propose QERA, a Quantitative Ethical Risk Assessment framework for systematically evaluating the ethical vulnerabilities of pathology foundation models (PFMs), as illustrated in Fig. \ref{fig:fig1}. Specifically, QERA assesses: (1) the extent to which patient-sensitive attributes, such as age, gender, or race, can be inferred from model representations, posing privacy concerns; (2) the degree to which non-diagnostic features influence downstream models, potentially undermining generalization and clinical reliability, and (3) performance disparities across demographic and institutional subgroups, which may exacerbate existing healthcare inequities. Beyond merely identifying these risks, we further investigate their underlying causes, revealing how current model training strategies result in these issues. Finally, we offer actionable insights into mitigating these ethical risks and empirically validate the effectiveness of our proposed solutions. The contributions and significances of this study can be summarized as follows:

\begin{itemize}
    \item \textbf{Revealing and Assessing Ethical Risks of PFMs}: We reveal that current PFMs may pose critical ethical risks and propose the first systematic and quantitative framework to assess them. Unlike prior evaluations that primarily focus on diagnostic accuracy or task generalizability, our framework explicitly targets ethical concerns that are essential for the safe and responsible deployment of PFMs in clinical practice.
    \item \textbf{Quantitative Evaluation Approach}: For different aspects of ethical risk, we propose concrete computational methods and conduct extensive experimental analyses. In contrast to previous studies that focus on theoretical discussions or qualitative summaries, our approach offers the first quantitative solution for ethical risk assessment, enabling results that are comparable, reproducible, and auditable.
    \item \textbf{Clinically Critical Ethical Perspectives}: Our framework integrates ethical dimensions that are of direct clinical relevance, including the potential leakage of patient-sensitive attributes, performance disparities across demographic and institutional subgroups, and the influence of diagnostically irrelevant features on model decisions. These correspond to key concerns in clinical deployment: privacy, fairness, and reliability.
    \item \textbf{Understanding and Mitigating the Origins of Ethical Risks}: We further investigate the underlying causes of such ethical risks, provide simple potential mitigation strategies, and empirically validate their effectiveness. These findings provide practical guidance for building future PFMs that are both more robust and ethically aligned.
    \item \textbf{Open-Access Evaluation Toolkit}: The proposed ethical risk evaluation framework will be released as an online platform and toolkit. This resource is intended to assist researchers and practitioners in rigorously assessing and facilitating the safe and ethical deployment of PFMs.
\end{itemize}

\section{Results}\label{result}
\subsection{The Overview}\label{subsec2}
In this study, we introduced the concept of quantitative ethical risk assessment and proposed the first quantitative framework for assessing the ethical risks of PFMs, as illustrated in Figure \ref{fig:fig1}. It focuses on three key dimensions of ethical concern. First, we assess privacy leakage, measuring the extent to which patient-sensitive attributes can be inferred from feature embeddings. Then we investigate clinical reliability, specifically the reliance on non-diagnostic features, as evidenced by performance degradation under out-of-distribution (OOD) setting. In addition, we evaluate group fairness by examining performance disparities across demographic and institutional subgroups. Our evaluation pipeline leverages over 20 curated datasets spanning a diverse range of organs, including the esophagus, lung, kidney, breast, colorectum, glioma, soft tissue, and uterus, as illustrated in Figure \ref{fig:fig1}d and detailed in Extended Table \ref{tab:dataset}. These evaluations provide a comprehensive and nuanced understanding of the ethical vulnerabilities inherent in current PFMs.
Additionally, to better understand the origins of these ethical risks, we investigated their underlying causes and validated our hypotheses through empirical experiments. We also offered potential insights into mitigating these issues and conducted preliminary experiments to explore their effectiveness.

We conducted evaluations on three representative state-of-the-art PFMs, namely UNI, Prov-GigaPath and Virchow, to demonstrate the applicability and effectiveness of our framework. These experiments span a diverse range of clinically relevant tasks, including patch classification, patch retrieval, WSI classification, and survival analysis, which correspond to real-world clinical applications such as cancer subtype classification, gene mutation prediction, prognosis estimation, and similar case retrieval.

\subsection{Privacy Leakage: Leakage of Patient-Sensitive Information}
Pathology images are processed by PFMs to generate fixed-dimensional feature embeddings, which are then utilized for various downstream clinical tasks. In this section, we examine whether these embeddings inadvertently encode patient-sensitive information. Specifically, we focus on two representative types of privacy related attributes, demographic information and medical institution, and assess the extent to which these can be inferred from the feature embedding.

\subsubsection{Feature embeddings encode patient demographic information}
We investigate whether PFMs encode patient demographic attributes, which are among the most common forms of sensitive personal information in clinical data. Specifically, we focused on three representative attributes: gender, race, and age. Based on the feature embeddings from PFMs, we trained simple classifiers to predict each attribute. The gender prediction task was formulated as a binary classification problem. For age, we divided patients into four age groups, forming a four-class classification task. Race prediction was also formulated as a binary classification task, including only White and Black or African American patients, as the number of samples from other racial groups was insufficient.
As shown in Figure \ref{fig:fig2}a–c and Extended Table \ref{tab:demographic_cls}, the classifiers achieved non-trivial prediction accuracy across all three demographic attributes. While the performance is not perfect, it consistently exceeds random chance, demonstrating that the demographic information is at least partially retained in the PFM-generated embeddings. These results suggest that PFMs may inadvertently encode patient demographic attributes, raising potential privacy concerns related to unintended information leakage.

\subsubsection{Feature embeddings encode patient’s medical institution}
\begin{figure}[htp]
\centering
\includegraphics[width=0.95\textwidth]{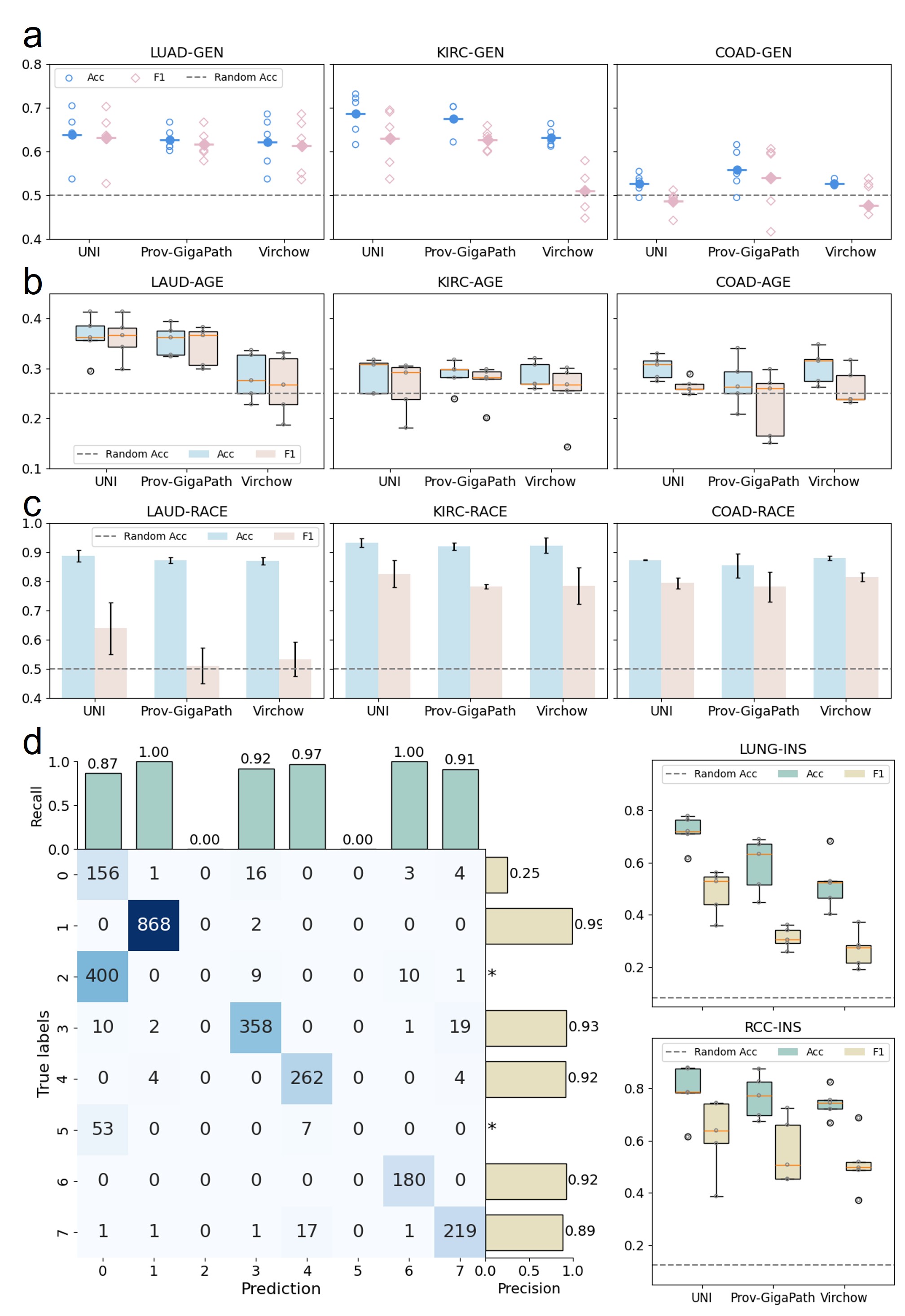}
\caption{Assessment of patient-sensitive information leakage. Performance of simple classifiers trained on PFM-derived feature embeddings to infer patient-sensitive attributes: (a) gender, (b) race, (c) age group, and (d) medical institution.}
\label{fig:fig2}
\end{figure}

Similarly, we assess whether the feature embeddings generated by PFMs also encode information related to medical institutions. Specifically, we used the TCGA-UT dataset \cite{komura2020histology}, a large-scale collection of pathology patches derived from human cancer tissues. From this dataset, we selected patches related to lung cancer, kidney cancer, breast cancer and colorectal cancer and used their corresponding institutions as labels to construct four datasets (LUNG-INS, RCC-INS, BRCA-INS, and COAD-INS) for evaluating institution classification performance. As shown in Figure \ref{fig:fig2}d and Extended Table \ref{tab:ins_cls}, by training a simple classifier to predict the institution, we observe high classification accuracy, indicating that institutional identity is indeed captured in the embedding space. This suggests that PFMs inadvertently encode non-clinical attributes related to institutions. These findings indicate that PFMs do not disregard non-diagnostic features but instead encode them, thereby posing a persistent risk of privacy leakage.

\subsection{Clinical Reliability: Impact of Non-Diagnostic Features}

\begin{figure}[htp]
\centering
\includegraphics[width=0.92\textwidth]{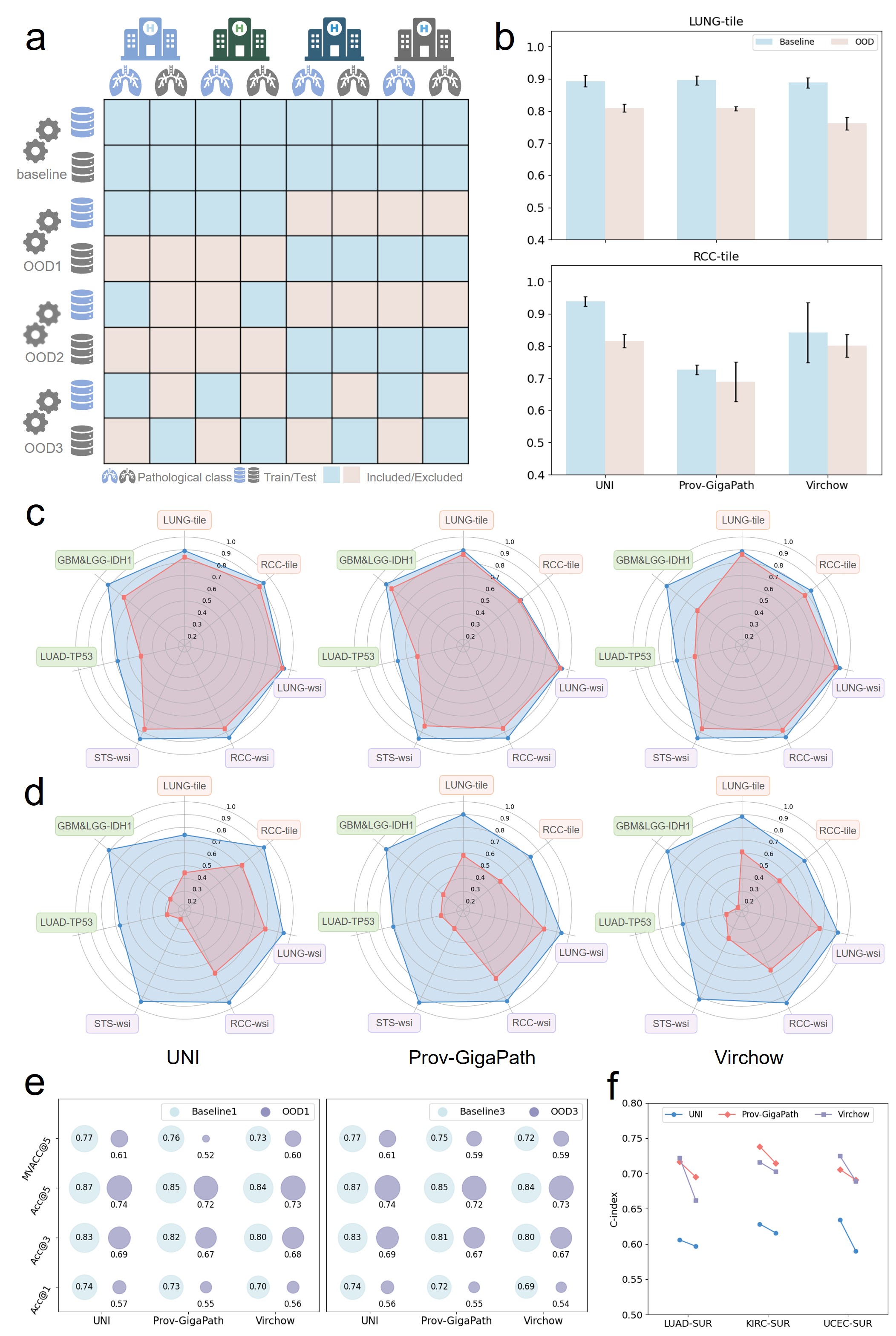}
\caption{Clinical Reliability Evaluation. \textbf{a.} Illustration of the three OOD settings designed to simulate various clinical deployment scenarios. \textbf{b.} Performance of two patch-level subtype classification datasets under OOD2 setting. \textbf{c-d.}, Model performance on seven classification datasets under OOD1 and OOD3 settings, including subtype classification and mutation prediction. \textbf{e.} Retrieval performance under OOD1 and OOD3 settings. \textbf{f.} Survival prediction performance on three datasets under OOD1.}
\label{fig:fig3}
\end{figure}

Since PFMs encode patient-sensitive features, these representations not only pose privacy risks but also raise concerns about clinical reliability. Such attributes are non-diagnostic in nature and may interfere with the training of downstream models. If these features are spuriously correlated with pathological labels, downstream models may inadvertently exploit them for diagnostic decision-making. These spurious correlations fail to reflect genuine medical knowledge and often lack consistency across multiple datasets, ultimately undermining model performance in practical clinical scenarios and raising concerns about its clinical trustworthiness.

In this section, we focus on the most easily inferred patient-sensitive attribute, the patient’s medical institution, to examine how such non-diagnostic features affect the generalizability of downstream models, thereby assessing the clinical reliability of PFMs. Specifically, we assess how downstream models perform in OOD settings. To this end, we conduct extensive experiments on common pathology analysis tasks, including patch classification, WSI classification, and image retrieval. For each task, the performance in the in-distribution (ID) test scenario serves as the baseline, and we examine the extent of performance degradation across OOD settings. 

As illustrated in Figure \ref{fig:fig3}a, we carefully design three types of OOD settings to reflect varying degrees of correlation between pathological labels and institution labels, simulating clinical scenarios with different levels of complexity and challenge.
The first setting (OOD1) involves training and test samples originating from different institutions, representing a common challenge encountered in clinical research. 
In the second setting (OOD2), training samples are sourced from multiple institutions, but each institution contributes samples from only a single class (e.g., Institution A provides only Class 0 samples, while Institution B provides only Class 1 samples). This setup simulates a scenario where institution-specific information exhibits spurious correlations with pathological labels. However, the test set is uniformly drawn from other institutions. 
The third setting (OOD3) shares the same training set as the second, but in the test set, each institution again provides only a single class of samples, with the institution-class correspondence completely reversed. In other words, the spurious correlations present in the training set are deliberately inverted in the test set. For example, if in the training set, Institution A provides only Class 0 samples and Institution B provides only Class 1 samples, then in the test set, Institution A provides only Class 1 samples and Institution B provides only Class 0 samples.

\subsubsection{Patch classification performance declines in OOD settings} 
Patch classification refers to the task of classifying image patches cropped from WSIs into different classes, such as normal, benign, or malignant tissue. Patch classification plays a crucial role in various pathology applications, including automated diagnosis, biomarker discovery, and aiding pathologists in identifying regions of interest. 

In this section, we selected patches related to lung and kidney tissues from the TCGA-UT dataset while excluding institutions with a limited number of patch images. This led to the creation of two new datasets for cancer subtype classification, LUNG-tile and RCC-tile. Patches were input into three PFMs to obtain the corresponding feature embeddings, which are then used to train classification models. Through these experiments, we systematically evaluate the impact of institution-specific information on patch classification tasks. 

Figure \ref{fig:fig3}b-d, Extended Table \ref{tab:ood_patch_lung} and Extended Table \ref{tab:ood_patch_rcc} present the performance of two datasets under the three OOD settings.  
The results reveal a significant decline in patch classification performance across all three OOD settings. As the spurious correlation between pathological labels and institution labels becomes progressively stronger across the OOD scenarios, the extent of performance degradation also increases.

\subsubsection{WSI prediction performance declines in OOD settings}
WSI analysis refers to the task of assigning a diagnostic or prognostic prediction to an entire WSI, such as distinguishing between different cancer subtypes or grading tumor severity. It is essential for automated pathology diagnosis and treatment decision support.

In this section, we selected multiple WSI analysis datasets covering various tasks, including cancer subtype classification (LUNG-wsi, RCC-wsi, SARC-wsi), gene mutation prediction (LUAD-TP53, GBM\&LGG-IDH1), and survival prediction (LUAD-SUR, KIRC-SUR, UCEC-SUR). Similarly, we employed predefined OOD settings to split the training and test sets. Notably, due to the limited number of WSIs, constructing the OOD2 setting was not feasible for the subtype classification task. Thus, only OOD1 and OOD3 settings were evaluated. For survival analysis tasks, which do not involve categorical labels, neither OOD2 nor OOD3 could be constructed, and evaluations were limited to the OOD1 setting. For all WSI-level tasks, we employed ABMIL \cite{ilse2018attention}, a widely used model in WSI analysis.

These results are presented in Figure \ref{fig:fig3}c-f, Extended Table \ref{tab:ood_wsi_subtype}, \ref{tab:ood_wsi_gene} and \ref{tab:ood_wsi_sur}. In most cases, downstream task models exhibit an obvious performance decline under OOD settings, with the degradation being particularly severe in the OOD3 setting, where stronger correlations between pathological labels and institutional identities are introduced. This phenomenon is consistently observed across subtype classification, mutation prediction, and survival analysis tasks, suggesting that diagnostic-irrelevant features, exemplified by institution-specific information, can interfere with model training and undermine generalizability.

In WSI-based analysis, cancer subtype classification and survival analysis are relatively common and comparatively straightforward tasks. However, existing PFMs still capture institution-specific information when extracting features, resulting in noticeable performance degradation under OOD settings.
Predicting gene mutation status from WSIs is a more challenging task, and the performance gap between ID and OOD scenarios is even more pronounced. This suggests that downstream models may be relying on spurious correlations, raising concerns that the performance of existing PFMs on complex tasks like gene mutation prediction may be overestimated.

\subsubsection{Image retrieval performance declines in OOD settings}
Pathology image retrieval refers to the task of searching for and retrieving similar WSIs or image patches from a large database based on a given query image. It enables pathologists to compare a query case with archived cases, aiding in more accurate and consistent diagnoses.

We conducted image retrieval experiments using the TCGA-Pan-Cancer dataset \cite{komura2022universal} and the experimental setup from UNI. Similarly, we evaluated retrieval performance under OOD scenarios by using feature embeddings extracted from PFMs. Notably, limited by institutional distribution of samples in this dataset, the OOD2 setting could not be constructed, and experiments were conducted only on OOD1 and OOD3. As shown in Figure \ref{fig:fig3}d and Extended Table \ref{tab:ood_retri}, image retrieval performance significantly degraded under OOD settings, indicating that institution-specific information also negatively impacts retrieval tasks.

\subsection{Fairness: Performance Disparities Across Multiple Groups}
Group fairness refers to the principle that models should maintain minimal performance disparities across different demographic or institutional subgroups. In the context of clinical AI, achieving group fairness is essential to ensure that diagnostic tools do not systematically disadvantage certain populations, which is crucial for promoting equitable healthcare delivery. To evaluate group fairness,  we adopted a standard downstream evaluation pipeline: feature embeddings were first extracted using PFMs, and then used to train task-specific diagnostic models. We then evaluated model performance across subgroups defined by patient gender, race, and medical institution. Experiments were conducted on six datasets representing diverse clinical scenarios.

\begin{figure}[htp]
\centering
\includegraphics[width=1.0\textwidth]{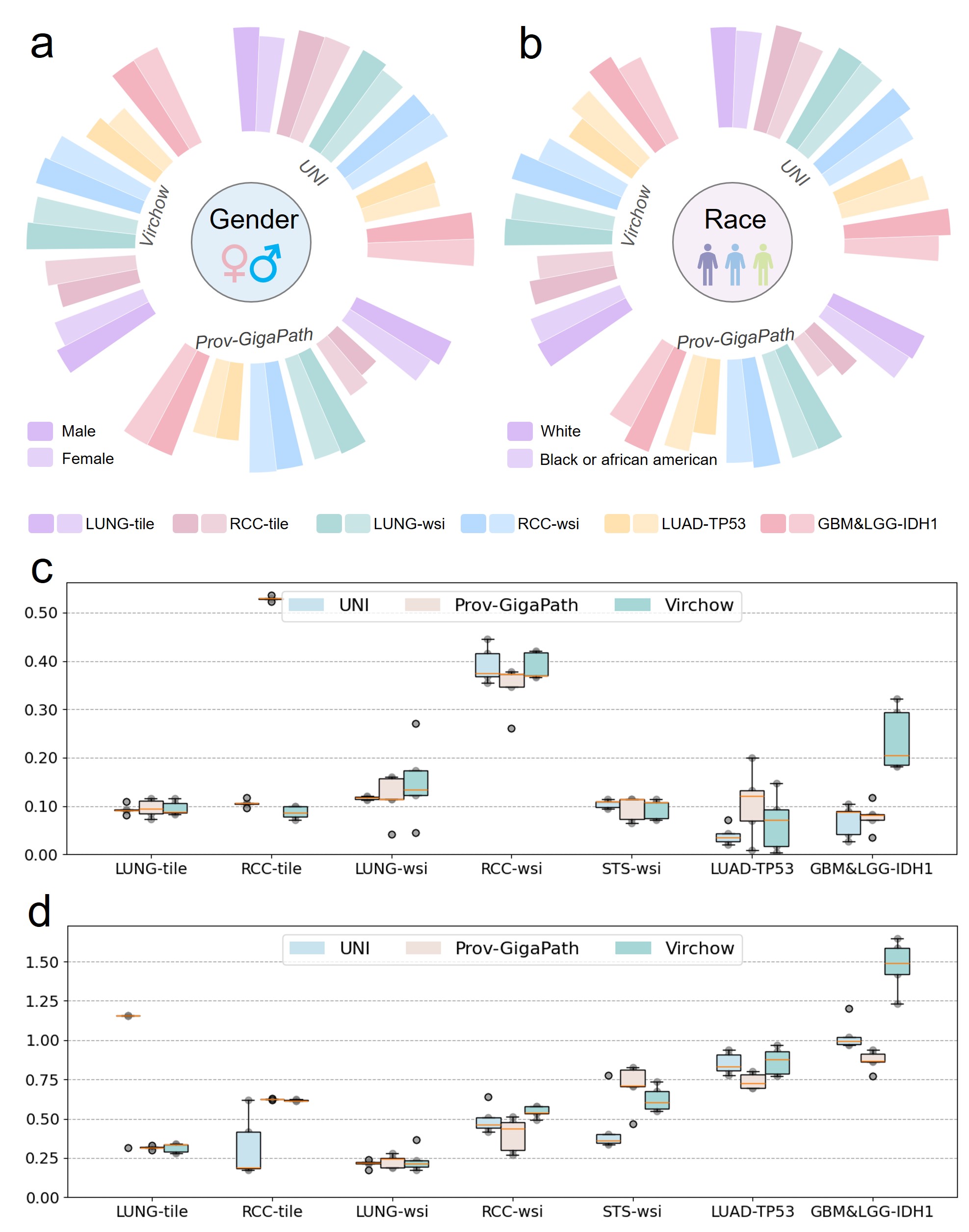}
\caption{Group Fairness Assessment. \textbf{a-b.} Accuracy differences across gender and racial subgroups on six datasets, where smaller gaps indicate better fairness. \textbf{c-d.} Coefficient of variation (CV) of diagnostic accuracy across institutions under OOD1 and OOD3 settings on six datasets. Lower CV values suggest better institutional fairness.}
\label{fig:fig4}
\end{figure}

Figure \ref{fig:fig4}a, Extended Table \ref{tab:fairness_gender_patch}, \ref{tab:fairness_gender_wsi_subtype} and \ref{tab:fairness_gender_wsi_gene} present the performance disparities observed when stratifying patients by gender, while Figure \ref{fig:fig4}b, Extended Table \ref{tab:fairness_race_patch}, \ref{tab:fairness_race_wsi_subtype} and \ref{tab:fairness_race_wsi_gene} show the corresponding results for patient race. In several cases, we observed considerable performance gaps between subgroups, suggesting potential fairness concerns. Since the number of institutions is not fixed, we directly computed the diagnostic accuracy for each individual hospital and then calculated the coefficient of variation (CV) across all institutions to quantify fairness. As shown in Figures \ref{fig:fig4}c-d, Extended Table \ref{tab:fairness_ins_ood1} and \ref{tab:fairness_ins_ood3}, the CV values are consistently greater than zero, where a value of zero would indicate perfect fairness, highlighting performance inconsistency across institutions. These findings suggest that downstream models trained on PFM-derived features may exhibit fairness issues, raising concerns about their equitable applicability in real-world clinical environments.

\subsection{Analysis of PFMs encoding non-diagnostic features}
As the medical institution of a patient is among the most readily inferred attributes from pathology images, we selected it as a representative non-diagnostic features to analyze the behavior of PFMs in encoding institution-specific information.

\subsubsection{Stain normalization fails to prevent encoding such features}
Institution-specific features in pathology images are often attributed to variations in staining protocols and scanner settings across institutions. Thus, stain normalization is commonly applied as a preprocessing technique to mitigate such non-diagnostic features. To investigate whether such normalization can effectively remove institution-specific signals, we applied stain normalization prior to feature extraction and compared the performance of institution classification before and after stain normalization. As shown in Figure \ref{fig:fig5}a and Extended Table \ref{tab:ins_cls}, although applying stain normalization reduces the institution classification accuracy, it does not eliminate the signal entirely. This indicates that common stain normalization cannot fully prevent PFMs from encoding such features, raising important questions about how to better constrain PFMs to focus on diagnosis-relevant information.

\subsubsection{Stain normalization fails to address the performance gap}
Staining differences are a key contributor to institution-specific features. As such, stain normalization has been widely adopted in prior studies as a common strategy to mitigate these biases. To evaluate whether stain normalization can help prevent PFMs from encoding diagnostic-irrelevant information, we conducted experiments to assess its impact on the generalization and robustness of downstream models. As shown in Figure \ref{fig:fig5}b and Extended Table \ref{tab:ood_stain_norm}, stain normalization leads to improved classification performance in only a subset of OOD settings, and the improvements are generally limited. These results suggest that while helpful in some cases, stain normalization alone is insufficient to fully eliminate institution-specific features from pathology images.

\subsubsection{Patient-sensitive features are learnable through pretraining}
In this section, we further investigate whether such institution-specific features can be acquired during pretraining and generalize across datasets and tasks like PFMs.
Specifically, we employed the ResNet and the TCGA-UT dataset, using the institution name of each image as the label to perform supervised pretraining, aiming to train a simple institution feature foundation model (IFFM). The dataset consists of pathology images from 32 cancer types, among which images from 27 cancer types were split into training, validation, and test sets (i.e., IFFM-pretrain) for pretraining and validating the institution information extraction model. Additionally, five cancer types related to the lung and kidney (i.e., LUNG-INS and RCC-INS) were designated as test sets. Pretraining such an IFFM is relatively straightforward, as evidenced by the obvious decline in loss during training, as shown in Figure \ref{fig:fig5}c.

Using this IFFM, we extracted institution-specific features from images for institution classification. Experiments were conducted on the LUNG-INS, RCC-INS, and OA-INS datasets. Here, OA is a dataset originally used for tumor detection and histological regression grading in oesophageal adenocarcinomas \cite{tolkach2023artificial}. We replaced the pathological labels with institution labels, creating the OA-INS dataset for institution classification. As shown in Figure \ref{fig:fig5}d, pretained IFFM successfully captures institution-specific features, achieving high classification performance. Notably, images from TCGA were excluded from the OA-INS dataset. Therefore, compared to the dataset used for pretraining, OA-INS serves as an external validation set. The IFFM still demonstrated excellent institution classification performance on this dataset. These results suggest that institution-specific features can be effectively extracted and generalized.

 \subsubsection{Why PFMs capture such non-diagnostic features}

In this section, we explore why PFMs capture non-diagnostic information from images and provide a simple validation of this phenomenon. One possible reason is that most PFMs are pretrained using self-supervised learning, a paradigm that tends to capture all differences between images rather than focusing solely on diagnostically relevant features. In this study, three PFMs employ the classic self-supervised learning method DINOv2 \cite{oquab2023dinov2}.
To learn diverse and generalizable representations, PFMs typically require large-scale datasets for pretraining. However, the limited data available from any single institution often necessitates the use of multi-institution datasets, which increases the risk of PFMs capturing institution-specific features. Similarly, these datasets often include samples from patients of varying genders and races, further introducing potential sources of non-diagnostic variability.

To validate this hypothesis, we conducted a set of experiments. Specifically, we trained two small-scale foundation models (VIT-S-SSL and VIT-S-FSL) using the same training data and model architecture but with different learning paradigms, self-supervised and fully supervised learning. Both models adopt the ViT-S architecture and are trained on the TCGA-UT dataset, where images from 27 cancer types are used for training, while the remaining 5 cancer types serve as the test set. The self-supervised model (VIT-S-SSL) is trained using the DINOv2 method, whereas the fully supervised model (VIT-S-FSL) is trained with the cancer-type labels provided by TCGA-UT. 

\begin{figure}[t]
\centering
\includegraphics[width=1.0\textwidth]{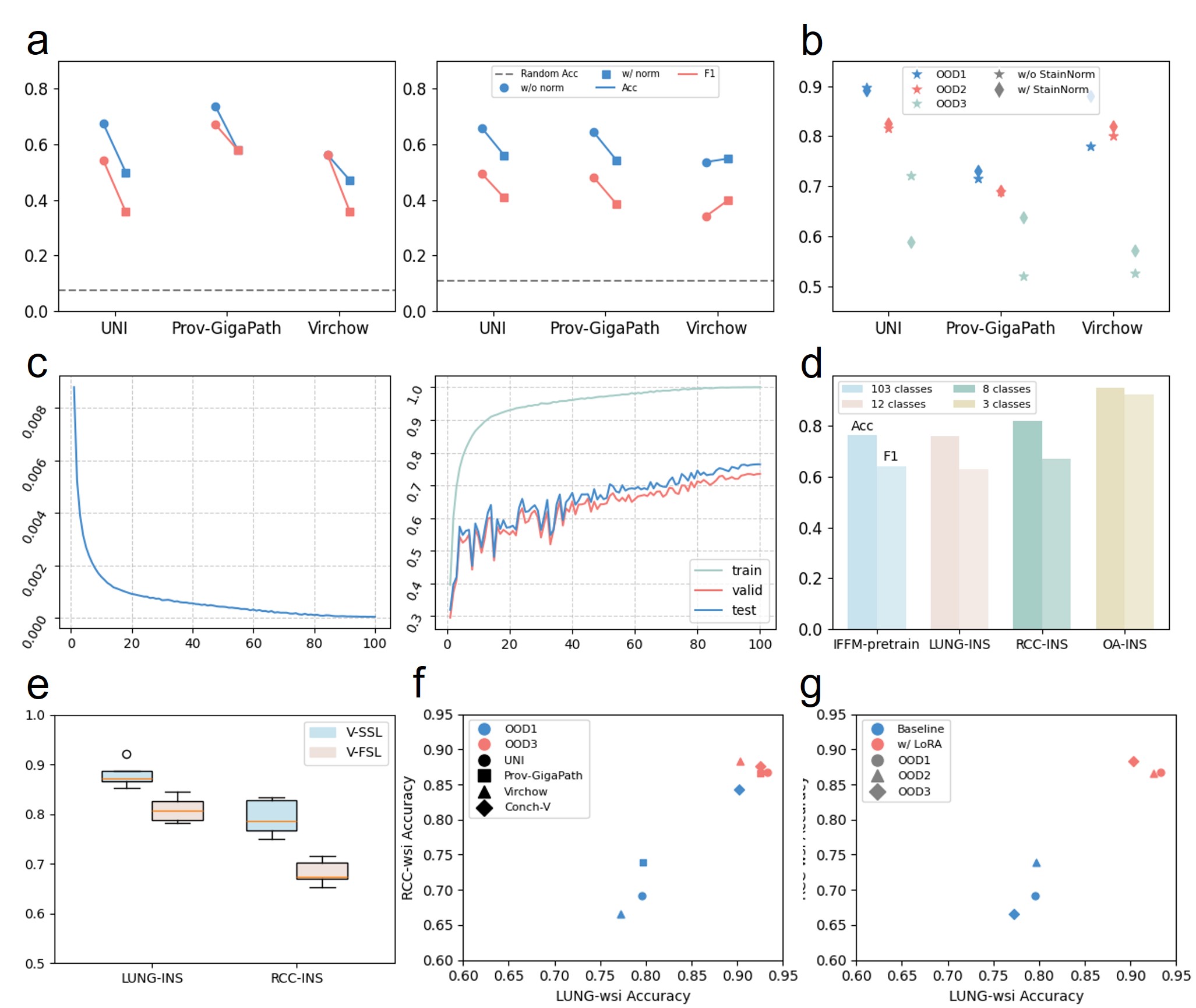}
\caption{Analysis of non-diagnostic features encoding in PFMs. \textbf{a.} Impact of stain normalization on institutional classification accuracy (BRCA-INS and COAD-INS). Dots on the left and right represent performance without and with stain normalization, respectively. \textbf{b.} Impact of stain normalization on performance degradation under OOD settings (the RCC-tile dataset). \textbf{c.} Training loss and accuracy curves of the IFFM model during supervised pretraining on institutional classification. \textbf{d.} Institution classification accuracy using institution-specific features extracted by the pretrained IFFM model. \textbf{e.} Comparison of institution classification accuracy using features from models pretrained with self-supervised vs. fully supervised learning. \textbf{f.} Performance comparison between CONCH-V and other PFMs under OOD settings. \textbf{g.} Impact of supervised fine-tuning (LoRA) of PFMs on downstream task performance.}
\label{fig:fig5}
\end{figure}

We then evaluated the institution classification performance of these two models using the LUNG-INS and RCC-INS datasets. As shown in Figure \ref{fig:fig5}e, VIT-S-SSL captured more institution-specific features, achieving higher institution classification accuracy. In contrast, the fully supervised model (VIT-S-FSL), guided by cancer-type labels during training, exhibited a reduced tendency to focus on institution-specific features. 

On the other hand, we introduced CONCH, a vision-language PFM designed for CPath. Unlike UNI, Prov-GigaPath, and Virchow, CONCH is trained by optimizing image-text matching loss rather than using a self-supervised approach. Its text branch serves as a supervised signal for the visual branch (Conch-V), making the visual encoder functionally equivalent to a vision foundation model trained in a supervised manner. To further investigate the reasons why PFMs capture institution-specific information, we evaluated Conch-V on institution classification and WSI classification in OOD settings, comparing its performance against the three previously mentioned PFMs. In most cases, using CONCH-V as the feature extractor results in poorer institution classification performance (Extended Table \ref{tab:ins_cls_ssl}), indicating that its feature embeddings indeed contain less institution-specific information. Figure \ref{fig:fig5}f, Extended Table \ref{tab:conch_lung} and \ref{tab:conch_rcc} presents the WSI classification performance. When using feature embeddings from CONCH-V, the downstream model maintains good performance in OOD3 settings, with only a slight decline compared to the corresponding baseline. In contrast, with the other three PFMs, the downstream models exhibit a significant performance decline in OOD3 settings. This suggests that CONCH-V is indeed less affected by the spurious correlation between institution-specific information and pathological labels.

\subsubsection{Supervised learning may help mitigate ethical risks}
In the previous section, we observed that feature extractors trained using fully supervised learning captured fewer institution-specific features compared to those trained with self-supervised learning. The vision encoder in CONCH, which is trained with the guidance of pathological knowledge, demonstrates better performance in OOD setting. These results suggest that incorporating appropriate supervision during finetuning may be beneficial for adapting PFMs to clinical downstream tasks. Given the large number of parameters in PFMs, we employed LoRA \cite{hu2021lora} for efficient fine-tuning. We evaluated this approach on the patch classification task, evaluating its performance in OOD setting. As shown in Figure \ref{fig:fig5}g and Extended Table \ref{tab:lora}, patch classification models with the finetuned PFM demonstrated improved performance across all three OOD settings on both datasets, especially in OOD3, where spurious correlations were the most severe. This suggests that applying appropriate supervision, whether during PFM pretraining or downstream model training, can help mitigate the influence of non-diagnostic features, thereby enhancing generalizability and robustness.

\section{Discussion}\label{sec12}
\subsection{Existing PFMs still perform well in various tasks}
Despite encoding non-diagnostic features, which may lead to potential privacy leakage, clinical reliability issues, and group fairness concerns, existing PFMs still demonstrate strong performance across a wide range of pathology image analysis tasks. This highlights their substantial advancement over earlier feature extractors pretrained on natural images.

Regarding privacy leakage, although the accuracy of predicting patient-sensitive information from PFM-generated feature embeddings generally exceeds random guessing, in most cases the accuracy remains limited. This indicates that while patient privacy information may be encoded, actual leakage remains difficult to achieve. In terms of group fairness, some tests show differences in diagnostic accuracy across groups, suggesting that PFMs have not completely excluded patient group information and still encode it to some extent. However, in many cases, the differences in diagnostic accuracy among groups are small and do not necessarily indicate serious healthcare fairness issues. Our evaluation of these two aspects focuses more on revealing potential risks rather than on problems that are already clearly manifested.

Regarding clinic reliability, we designed three OOD settings to examine how downstream models are affected by spurious correlations between non-diagnostic features and pathology labels. Among them, OOD1 represents the most likely scenario in real clinical settings, where the performance of downstream models declines slightly but remains effective, demonstrating the practicality of current PFMs. In contrast, OOD2 and OOD3 serve as rigorous test scenarios to evaluate the performance of PFMs' feature embeddings under extreme conditions. Ideally, PFMs should maintain strong performance even in these settings. However, if performance degradation occurs, it does not imply that PFMs are unusable. Instead, it suggests that PFMs capture both diagnostically relevant pathology patterns and  non-diagnostic features. In common clinical scenarios, pathology features dominate, allowing PFMs to remain broadly applicable to pathology analysis tasks. However, in extreme settings, the spurious correlation between non-diagnostic features and pathology labels becomes too strong, leading clinic reliability issues.


\subsection{Innovations in training strategies are worth considering}
The PFMs evaluated in this study were all trained using self-supervised learning, without explicitly provided supervision signals related to pathological knowledge. As a result, these PFMs were inherently encouraged to learn all differences between images, including both pathological patterns and non-diagnostic features. In section 2.5.4, we analyzed how self-supervised and fully supervised training methods influence the retention of non-diagnostic features. Additionally, we evaluated another vision-language PFM, CONCH, whose vision encoder (CONCH-V) can be considered a vision foundation model trained with supervision. We found that CONCH-V performed better in OOD settings compared to purely vision-based PFMs trained with self-supervised learning.
Similarly, in Section 2.5.5 , we found that finetuning pretrained PFMs with downstream task labels can also mitigate the spurious correlation between institution-specific features and pathological labels, thereby improving the performance of downstream models in OOD settings. 

Therefore, incorporating supervision or guidance related to pathological features during PFM training is meaningful. However, due to the ultra-high resolution of pathology images, obtaining fine-grained annotations is relatively challenging. Thus, fully supervised training is impractical. In conclusion, we encourage innovation in existing self-supervised algorithms by designing specific self-supervised signals for pathological features and non-diagnostic features to decouple the two or suppress non-diagnostic features.

\subsection{PFM performance in extreme scenarios requires attention}
To evaluate the impact of non-diagnostic information on downstream model performance, this study defines three OOD scenarios. OOD1 represents the most common setting, covering the majority of clinical diagnostic tasks. However, for rare diseases where data from a single institution may not encompass all categories, scenarios described by OOD2 and OOD3 could emerge. Therefore, to ensure that PFMs remain applicable to a wide range of pathology analysis tasks, the extreme conditions represented by OOD2 and OOD3 also warrant attention.

On the other hand, evaluating OOD performance in extreme scenarios helps determine whether a downstream model relies on spurious correlations for decision-making. When tackling challenging tasks, such as predicting gene mutations from WSI, the pathological features learned by PFMs may be subtle. In such cases, if non-diagnostic features are spuriously correlated with pathological labels, the downstream model may rely on these features for gene mutation prediction, leading to an overestimation of its performance. By testing under extreme OOD settings, where the likelihood of spurious correlations remaining consistent between the training and test sets is lower, we can identify such spurious correlations through a noticeable decline in model performance on OOD settings.

\subsection{Ethical risks start with data, but PFMs can do better}
It is worth noting that the ethical risks identified in our study may not solely stem from PFMs, but could also be influenced by biases in dataset composition. However, adjusting the sample distribution of training data is often impractical, as it requires substantial human effort, time, and domain expertise. For instance, rare disease cohorts are inherently limited in size, making it difficult to collect sufficient data, let alone ensure balanced subgroup representation. Moreover, many non-diagnostic features, such as subtle staining differences or institution-specific artifacts, are often visually imperceptible, making manual control even more challenging. Therefore, mitigating spurious correlations and associated ethical risks by manually adjusting downstream training distributions is difficult to achieve in practice.
In contrast, PFMs are pretrained on massive and diverse pathology datasets, and have likely been exposed to a wide spectrum of pathological patterns. Through learning numerous linear and nonlinear transformations, PFMs may capture sub-visual features beyond human perceptual capacity. This positions PFMs as promising candidates to address and reduce ethical risks in downstream models, if appropriately guided during their design and deployment.

\subsection{Promoting lab-to-clinic transition via ethical risk assessment}
Conducting ethical risk assessments for PFMs is critically important, especially as these models are increasingly deployed in clinical workflows. Unlike traditional models trained for narrow, well-defined tasks, PFMs are intended to serve as general-purpose feature extractors across diverse diagnostic and prognostic applications. This broad applicability magnifies their influence on clinical decisions, and thus their embedded biases and unintended behaviors can have far-reaching consequences. As demonstrated in this study, PFMs may inadvertently encode non-diagnostic features such as patient demographics or institution-specific features. This not only raises concerns about patient privacy, but also threatens the fairness and reliability of downstream diagnostic systems. Without systematic evaluation, these risks may remain hidden, undermining trust in clinical AI. Therefore, establishing a quantitative framework for assessing ethical risks, as proposed in this work, is essential for ensuring the safe, equitable, and responsible deployment of PFMs in real-world medical settings.

\section{Conclusion}
PFMs have demonstrated remarkable potential in advancing computational pathology by enabling feature extraction for a wide range of diagnostic tasks. However, as PFMs are increasingly considered for clinical deployment, concerns around their ethical risks—such as privacy leakage, compromised clinical reliability, and group fairness—must be carefully addressed. This highlights the urgent need for systematic ethical risk assessment prior to real-world application.

In this study, we proposed a quantitative framework for evaluating ethical risks in PFMs across three key dimensions: privacy, reliability, and fairness. We propose an assessment framework that systematically measures key dimensions of ethical concern: the degree to which patient-sensitive attributes can be inferred from model representations, the influence of non-diagnostic features on model decisions, and the extent of performance disparities across demographic and institutional subgroups. 

Our work introduces the concept of quantitative ethical risk assessment and proposes the first quantitative framework for assessing  ethical risks in PFMs. This allows for reproducible and auditable assessments of whether PFMs encode non-diagnostic and potentially harmful information. By establishing a practical and extensible evaluation framework, our aim is to promote the safe, fair, and reliable deployment of PFMs from research laboratories to clinical practice.

\section{Method}\label{method}
\subsection{The quantitative framework}
We propose a systematic and quantitative framework to assess the ethical risks of pathology foundation models (PFMs) from three critical perspectives: privacy leakage, clinical reliability, and group fairness. While previous discussions on ethical risks have largely remained at a conceptual or theoretical level, our approach introduces concrete, operational metrics for each type of risk. These metrics enable evaluations that are comparable, reproducible, and auditable, thereby offering a practical toolkit for identifying and monitoring ethical vulnerabilities in PFMs.

\subsubsection{Privacy Leakage Assessments} To evaluate the risk of privacy leakage in PFMs, we assess whether patient-sensitive attributes can be inferred from the feature embeddings generated by them. Specifically, we first use PFMs to extract features from pathology images and then train lightweight classifiers to predict demographic attributes such as gender, race, and age group, as well as the medical institution where the patient received care. For patch-level images, we employ a multilayer perceptron (MLP) as the classifier, while for WSIs, we adopt an MIL framework. Model performance is measured by classification accuracy and compared to random guessing baselines. For example, in the binary gender classification task, a random guess would yield an accuracy of 0.5. Higher accuracy indicates that more patient-specific information is embedded in the features extracted by the PFM, implying a greater risk of privacy leakage.

\subsubsection{Clinical Reliability Assessments}
To evaluate the clinical reliability of PFMs, we investigate how the presence of non-diagnostic features—specifically, patient-sensitive attributes encoded in feature embeddings—affects the robustness of downstream models under varying deployment scenarios. We focus on one of the most easily inferred patient attributes, the medical institution, to examine its potential confounding effect on downstream tasks. Given that spurious correlations may exist between institution labels and pathological labels, we meticulously design three types of OOD settings to simulate real-world clinical scenarios, as illustrated in Figure 3a.

\textbf{In OOD1}, samples in the training and testing sets come from different institutions. This simulates the most common deployment scenario in multi-center clinical studies, where external validation is used to assess generalizability. For example, the training set is from Institution A and Institution B, while the test set is from Institution C and Institution D.

\textbf{In OOD2}, each institution in the training set contributes samples from only a single diagnostic class, creating an artificial correlation between institutional identity and pathological label. In contrast, the test set consists of samples from entirely different institutions, each containing a balanced mix of diagnostic classes. This design breaks the spurious correlation at test time, allowing us to evaluate whether the model relies on such non-diagnostic features. For instance, in training set, Class 0 and Class 1 are from Institution A and Institution B, respectively. All samples are from Institution C and Institution D. 

\textbf{OOD3} further intensifies this discrepancy by inverting the institution-label correlation in the test set relative to the training set. For example, if Institution A only provides Class 0 samples and Institution B provides Class 1 in training, then in testing, Institution A will only provide Class 1 and Institution B will provide Class 0. This setup is intentionally strict: if a model relies entirely on spurious correlations, it may fail completely on the test set.

For each OOD setting, we construct a baseline by merging the samples from its training and test sets and then randomly splitting them into new training and test sets with balanced class distributions, without introducing any spurious correlations. Notably, each OOD scenario shares the exact same set of samples with its corresponding baseline, differing only in the splitting of training and test sets. Due to the stricter sample distribution requirements in the OOD3 setting, the number of included patches is slightly lower than in OOD1 and OOD2. Consequently, the baseline for OOD3 differs from those of OOD1 and OOD2. This baseline allows us to assess the extent to which performance degradation under OOD conditions is due to the model's reliance on non-diagnostic features. A robust and clinically reliable model should maintain stable performance across all OOD settings. Significant performance degradation, on the other hand, suggests that the feature embeddings are contaminated by non-diagnostic attributes, undermining their reliability in real clinical use.

\subsubsection{Group Fairness Assessments}
In the context of clinical AI, ensuring group fairness is essential to guarantee that diagnostic tools do not systematically disadvantage certain populations. To evaluate group fairness, feature embeddings were first extracted using PFMs, followed by training downstream models on these embeddings for various diagnostic tasks. We then assessed the performance gap across subgroups, such as comparing diagnostic accuracy between male and female patients. A larger performance gap suggests a model's bias toward certain subgroups, indicating potential fairness issues. Given that the number of institutions is not fixed, we computed the diagnostic accuracy for each individual hospital and then calculated the coefficient of variation (CV) across all institutions to quantify fairness. Ideally, the CV value should be as close to zero as possible, indicating minimal disparity in diagnostic accuracy across different institutions. A higher CV suggests that the model may exhibit unfair biases towards certain hospitals, pointing to potential fairness concerns in the model's decision-making process.

\subsection{Dataset curation}
In total, this study employs 26 curated datasets, which are derived from existing sources through sample selection or label adjustment to support various ethical risk evaluation tasks. Detailed dataset information is provided in Extended Table 1.

\subsubsection{Datasets for Privacy Leakage Assessments.} To evaluate the extent to which PFMs encode patient-sensitive information, we selected WSIs from the TCGA-LUAD, TCGA-KIRC, and TCGA-COAD cohorts. Demographic attributes, including age, gender, and race, were extracted from the corresponding clinical reports to serve as labels for privacy risk assessment. These datasets are summarized in Extended Table 1 (IDs 1–9). The TCGA-UT dataset is a large-scale collection of pathology patches derived from TCGA WSIs. We selected patches related to the lung, kidney, colorectal, and breast tissues, and retrieved the associated medical institution labels based on the corresponding WSI identifiers. To ensure data quality and representation, institutions with fewer than 300 patches were excluded. This curation process resulted in four datasets used to evaluate whether PFMs encode institutional identity (IDs 10–13). Similarly, we constructed a new dataset, denoted as IFFM-pretrain (ID 14 in Extended Table 1), using the entire set of pathology images in the TCGA-UT dataset along with their associated institution labels. This dataset was used to pretrain an institutional feature extraction model (IFFM). To evaluate the generalizability of the pretrained IFFM, we curated an additional dataset, OA-INS (ID 15), from the OA dataset by excluding patches originating from TCGA and assigning institution labels accordingly.

\subsubsection{Datasets for Fairness and Clinical Reliability Assessments.}
To evaluate the fairness and clinical reliability of PFMs, we curated datasets covering three major downstream tasks: patch classification, WSI-level analysis, and patch retrieval. 
For the patch classification task, we selected lung- and kidney-related patches from the TCGA-UT dataset and used their corresponding cancer subtypes as labels, resulting in two patch-level datasets: LUNG-tile and RCC-tile (IDs 16 and 17). 
For WSI-level subtype classification tasks, we curated three datasets, namely LUAD-wsi, RCC-wsi and STS-wsi. The LUNG-wsi dataset (ID 18) consists of TCGA-LUAD and TCGA-LUSC, corresponding to two common lung cancer subtypes, lung adenocarcinoma (LUAD) and lung squamous cell carcinoma (LUSC). The RCC-wsi dataset (ID 19) comprises TCGA-KIRC, TCGA-KIRP, and TCGA-KICH, representing three major kidney cancer subtypes, kidney renal clear cell carcinoma (KIRC), kidney renal papillary cell carcinoma (KIRP), and kidney chromophobe carcinoma (KICH). The STS-wsi dataset (ID 20) is private, including two common types of soft tissue sarcomas, liposarcoma and leiomyosarcoma. 
For the gene mutation prediction task, we conducted experiments on two datasets, LUAD-TP53 (ID 21) and GBM\&LGG-IDH1 (ID 22). The LUAD-TP53 dataset consists of WSIs from TCGA-LUAD, with labels indicating whether the TP53 gene is mutated. The GBM\&LGG-IDH1 dataset includes WSIs from TCGA-GBM and TCGA-LGG, with labels indicating the mutation status of the IDH1 gene. 
For survival analysis, we extracted survival data from clinical reports for TCGA-LUAD, TCGA-KIRC, and TCGA-UCEC, creating three corresponding datasets (IDs 23–25). 
For the patch retrieval experiment, we constructed the Pan-cancer-RETR dataset (ID 26) by sampling 300 patches from each of the 32 cancer types in TCGA-UT. If a cancer type contained fewer than 300 patches, all available patches were included.

\subsection{Experimental Setup}
\subsubsection{Patch classification tasks}
For patch classification tasks, all patch-level images were first processed by PFMs to extract feature embeddings, which were then fed into a lightweight classifier composed of two linear layers. All models were trained with cross-entropy loss using the Adam optimizer (initial learning rate = 5e-4) and cosine annealing (every 10 epochs). Each dataset was first randomly split using stratified sampling, with $80\%$ of the samples used for five-fold cross-validation and the remaining $20\%$ held out as a fixed test set. Each fold was trained for 50 epochs with a batch size of 256. All splits ensured strict slide-level separation to avoid data leakage. All experiments were conducted on a server with 8 NVIDIA RTX 3090 GPUs. 


\subsubsection{WSI classification tasks} 
For the WSI classification tasks, each whole-slide image was processed using PFMs to extract patch-level embeddings, which were subsequently aggregated using the Attention-Based Multiple Instance Learning (ABMIL) framework. All models were trained with cross-entropy loss using the Adam optimizer (initial learning rate = 2e-4) and cosine annealing (updated every 10 epochs). Each dataset was first randomly split using stratified sampling, with $80\%$ of the slides used for five-fold cross-validation and the remaining $20\%$ held out as a fixed test set. Each fold was trained for 50 epochs with a batch size of 8. All splits ensured strict patient-level separation to prevent data leakage. Experiments were conducted on an 8×NVIDIA RTX 3090 GPU server. 

For survival analysis tasks, due to the complexity of stratified sampling that involves controlling multiple variables such as risk group, censoring status, and institution labels, we directly adopted five-fold cross-validation. Given the characteristics of survival loss computation, we used a nominal batch size of 1 with 32-step gradient accumulation to obtain an effective batch size of 32. The model was optimized using the negative log-likelihood survival loss, along with $\ell_2$ weight decay ($\lambda = 10^{-5}$) and $\ell_1$ regularization ($\lambda = 10^{-4}$) to improve training stability and encourage sparsity in the learned features.

\subsubsection{Pretraining settings}
\textbf{Institution feature foundation model.} We employed the ResNet18 and conducted supervised pretraining on the TCGA-UT dataset, using the institution name as the label. This setup encourages the model to capture institution-specific information. The dataset comprises pathology images from 32 cancer types, of which 27 were used to construct training, validation, and test splits for model development and evaluation. The remaining five cancer types, related to the lung and kidney, were held out as independent test sets (LUNG-INS and RCC-INS) to assess the model’s generalization to unseen cancer types. The model was trained using cross-entropy loss with a batch size of 256, a maximum of 100 epochs, and an initial learning rate of 5e-4. The checkpoint with the highest validation accuracy was selected as the final model.

\textbf{VIT-S-SSL and VIT-S-FSL.} 
To examine whether the encoding of patient-sensitive information by PFMs is a consequence of self-supervised learning encouraging the capture of all type of features, we pretrained two lightweight feature extractors (ViT-S-SSL and ViT-S-FSL) using the same data but with different strategies. Specifically, we used pathology patches from 27 cancer types in the TCGA-UT dataset for pretraining and held out patches from the remaining 5 cancer types related to the lung and kidney as test data. For ViT-S-SSL, we ignored the cancer-type labels and applied self-supervised learning using the DINOv2 framework. In contrast, ViT-S-FSL was trained under a fully supervised setting using cancer-type labels and a cross-entropy loss. All other training configurations were kept identical across both models.

\subsubsection{PFMs fine-tuning}
To efficiently fine-tune the pretrained PFMs, we employed Low-Rank Adaptation (LoRA)—a parameter-efficient fine-tuning technique that inserts trainable low-rank matrices into the model while keeping the original weights frozen. In our experiments, the LoRA modules were configured with a rank $r=4$ and a scaling factor $\alpha = 4$. The finetuning was conducted using the Adam optimizer, a batch size of 32, and a learning rate of 5e-6 to ensure stability. Models were trained for 20 epochs, considering the relatively small size of the finetuning dataset.

\backmatter

\begin{appendices}

\end{appendices}


\bibliography{main}


\begin{thebibliography}{28}
\ifx \bisbn   \undefined \def \bisbn  #1{ISBN #1}\fi
\ifx \binits  \undefined \def \binits#1{#1}\fi
\ifx \bauthor  \undefined \def \bauthor#1{#1}\fi
\ifx \batitle  \undefined \def \batitle#1{#1}\fi
\ifx \bjtitle  \undefined \def \bjtitle#1{#1}\fi
\ifx \bvolume  \undefined \def \bvolume#1{\textbf{#1}}\fi
\ifx \byear  \undefined \def \byear#1{#1}\fi
\ifx \bissue  \undefined \def \bissue#1{#1}\fi
\ifx \bfpage  \undefined \def \bfpage#1{#1}\fi
\ifx \blpage  \undefined \def \blpage #1{#1}\fi
\ifx \burl  \undefined \def \burl#1{\textsf{#1}}\fi
\ifx \doiurl  \undefined \def \doiurl#1{\url{https://doi.org/#1}}\fi
\ifx \betal  \undefined \def \betal{\textit{et al.}}\fi
\ifx \binstitute  \undefined \def \binstitute#1{#1}\fi
\ifx \binstitutionaled  \undefined \def \binstitutionaled#1{#1}\fi
\ifx \bctitle  \undefined \def \bctitle#1{#1}\fi
\ifx \beditor  \undefined \def \beditor#1{#1}\fi
\ifx \bpublisher  \undefined \def \bpublisher#1{#1}\fi
\ifx \bbtitle  \undefined \def \bbtitle#1{#1}\fi
\ifx \bedition  \undefined \def \bedition#1{#1}\fi
\ifx \bseriesno  \undefined \def \bseriesno#1{#1}\fi
\ifx \blocation  \undefined \def \blocation#1{#1}\fi
\ifx \bsertitle  \undefined \def \bsertitle#1{#1}\fi
\ifx \bsnm \undefined \def \bsnm#1{#1}\fi
\ifx \bsuffix \undefined \def \bsuffix#1{#1}\fi
\ifx \bparticle \undefined \def \bparticle#1{#1}\fi
\ifx \barticle \undefined \def \barticle#1{#1}\fi
\bibcommenthead
\ifx \bconfdate \undefined \def \bconfdate #1{#1}\fi
\ifx \botherref \undefined \def \botherref #1{#1}\fi
\ifx \url \undefined \def \url#1{\textsf{#1}}\fi
\ifx \bchapter \undefined \def \bchapter#1{#1}\fi
\ifx \bbook \undefined \def \bbook#1{#1}\fi
\ifx \bcomment \undefined \def \bcomment#1{#1}\fi
\ifx \oauthor \undefined \def \oauthor#1{#1}\fi
\ifx \citeauthoryear \undefined \def \citeauthoryear#1{#1}\fi
\ifx \endbibitem  \undefined \def \endbibitem {}\fi
\ifx \bconflocation  \undefined \def \bconflocation#1{#1}\fi
\ifx \arxivurl  \undefined \def \arxivurl#1{\textsf{#1}}\fi
\csname PreBibitemsHook\endcsname

\bibitem[\protect\citeauthoryear{Song et~al.}{2023}]{song2023artificial}
\begin{barticle}
\bauthor{\bsnm{Song}, \binits{A.H.}},
\bauthor{\bsnm{Jaume}, \binits{G.}},
\bauthor{\bsnm{Williamson}, \binits{D.F.}},
\bauthor{\bsnm{Lu}, \binits{M.Y.}},
\bauthor{\bsnm{Vaidya}, \binits{A.}},
\bauthor{\bsnm{Miller}, \binits{T.R.}},
\bauthor{\bsnm{Mahmood}, \binits{F.}}:
\batitle{Artificial intelligence for digital and computational pathology}.
\bjtitle{Nature Reviews Bioengineering}
\bvolume{1}(\bissue{12}),
\bfpage{930}--\blpage{949}
(\byear{2023})
\end{barticle}
\endbibitem

\bibitem[\protect\citeauthoryear{Cui and Zhang}{2021}]{cui2021artificial}
\begin{barticle}
\bauthor{\bsnm{Cui}, \binits{M.}},
\bauthor{\bsnm{Zhang}, \binits{D.Y.}}:
\batitle{Artificial intelligence and computational pathology}.
\bjtitle{Laboratory Investigation}
\bvolume{101}(\bissue{4}),
\bfpage{412}--\blpage{422}
(\byear{2021})
\end{barticle}
\endbibitem

\bibitem[\protect\citeauthoryear{Quellec et~al.}{2017}]{quellec2017multiple}
\begin{barticle}
\bauthor{\bsnm{Quellec}, \binits{G.}},
\bauthor{\bsnm{Cazuguel}, \binits{G.}},
\bauthor{\bsnm{Cochener}, \binits{B.}},
\bauthor{\bsnm{Lamard}, \binits{M.}}:
\batitle{Multiple-instance learning for medical image and video analysis}.
\bjtitle{IEEE reviews in biomedical engineering}
\bvolume{10},
\bfpage{213}--\blpage{234}
(\byear{2017})
\end{barticle}
\endbibitem

\bibitem[\protect\citeauthoryear{Gadermayr and Tschuchnig}{2024}]{gadermayr2024multiple}
\begin{botherref}
\oauthor{\bsnm{Gadermayr}, \binits{M.}},
\oauthor{\bsnm{Tschuchnig}, \binits{M.}}:
Multiple instance learning for digital pathology: A review of the state-of-the-art, limitations \& future potential.
Computerized Medical Imaging and Graphics,
102337
(2024)
\end{botherref}
\endbibitem

\bibitem[\protect\citeauthoryear{Chanda et~al.}{2024}]{chanda2024new}
\begin{botherref}
\oauthor{\bsnm{Chanda}, \binits{D.}},
\oauthor{\bsnm{Aryal}, \binits{M.}},
\oauthor{\bsnm{Soltani}, \binits{N.Y.}},
\oauthor{\bsnm{Ganji}, \binits{M.}}:
A new era in computational pathology: A survey on foundation and vision-language models.
arXiv preprint arXiv:2408.14496
(2024)
\end{botherref}
\endbibitem

\bibitem[\protect\citeauthoryear{He et~al.}{2016}]{he2016deep}
\begin{bchapter}
\bauthor{\bsnm{He}, \binits{K.}},
\bauthor{\bsnm{Zhang}, \binits{X.}},
\bauthor{\bsnm{Ren}, \binits{S.}},
\bauthor{\bsnm{Sun}, \binits{J.}}:
\bctitle{Deep residual learning for image recognition}.
In: \bbtitle{Proceedings of the IEEE Conference on Computer Vision and Pattern Recognition},
pp. \bfpage{770}--\blpage{778}
(\byear{2016})
\end{bchapter}
\endbibitem

\bibitem[\protect\citeauthoryear{Huang et~al.}{2017}]{huang2017densely}
\begin{bchapter}
\bauthor{\bsnm{Huang}, \binits{G.}},
\bauthor{\bsnm{Liu}, \binits{Z.}},
\bauthor{\bsnm{Van Der~Maaten}, \binits{L.}},
\bauthor{\bsnm{Weinberger}, \binits{K.Q.}}:
\bctitle{Densely connected convolutional networks}.
In: \bbtitle{Proceedings of the IEEE Conference on Computer Vision and Pattern Recognition},
pp. \bfpage{4700}--\blpage{4708}
(\byear{2017})
\end{bchapter}
\endbibitem

\bibitem[\protect\citeauthoryear{Wang et~al.}{2022}]{wang2022transformer}
\begin{barticle}
\bauthor{\bsnm{Wang}, \binits{X.}},
\bauthor{\bsnm{Yang}, \binits{S.}},
\bauthor{\bsnm{Zhang}, \binits{J.}},
\bauthor{\bsnm{Wang}, \binits{M.}},
\bauthor{\bsnm{Zhang}, \binits{J.}},
\bauthor{\bsnm{Yang}, \binits{W.}},
\bauthor{\bsnm{Huang}, \binits{J.}},
\bauthor{\bsnm{Han}, \binits{X.}}:
\batitle{Transformer-based unsupervised contrastive learning for histopathological image classification}.
\bjtitle{Medical image analysis}
\bvolume{81},
\bfpage{102559}
(\byear{2022})
\end{barticle}
\endbibitem

\bibitem[\protect\citeauthoryear{Chen et~al.}{2024}]{chen2024towards}
\begin{barticle}
\bauthor{\bsnm{Chen}, \binits{R.J.}},
\bauthor{\bsnm{Ding}, \binits{T.}},
\bauthor{\bsnm{Lu}, \binits{M.Y.}},
\bauthor{\bsnm{Williamson}, \binits{D.F.}},
\bauthor{\bsnm{Jaume}, \binits{G.}},
\bauthor{\bsnm{Song}, \binits{A.H.}},
\bauthor{\bsnm{Chen}, \binits{B.}},
\bauthor{\bsnm{Zhang}, \binits{A.}},
\bauthor{\bsnm{Shao}, \binits{D.}},
\bauthor{\bsnm{Shaban}, \binits{M.}}, \betal:
\batitle{Towards a general-purpose foundation model for computational pathology}.
\bjtitle{Nature Medicine}
\bvolume{30}(\bissue{3}),
\bfpage{850}--\blpage{862}
(\byear{2024})
\end{barticle}
\endbibitem

\bibitem[\protect\citeauthoryear{Xu et~al.}{2024}]{xu2024whole}
\begin{botherref}
\oauthor{\bsnm{Xu}, \binits{H.}},
\oauthor{\bsnm{Usuyama}, \binits{N.}},
\oauthor{\bsnm{Bagga}, \binits{J.}},
\oauthor{\bsnm{Zhang}, \binits{S.}},
\oauthor{\bsnm{Rao}, \binits{R.}},
\oauthor{\bsnm{Naumann}, \binits{T.}},
\oauthor{\bsnm{Wong}, \binits{C.}},
\oauthor{\bsnm{Gero}, \binits{Z.}},
\oauthor{\bsnm{Gonz{\'a}lez}, \binits{J.}},
\oauthor{\bsnm{Gu}, \binits{Y.}}, et al.:
A whole-slide foundation model for digital pathology from real-world data.
Nature,
1--8
(2024)
\end{botherref}
\endbibitem

\bibitem[\protect\citeauthoryear{Vorontsov et~al.}{2024}]{vorontsov2024foundation}
\begin{barticle}
\bauthor{\bsnm{Vorontsov}, \binits{E.}},
\bauthor{\bsnm{Bozkurt}, \binits{A.}},
\bauthor{\bsnm{Casson}, \binits{A.}},
\bauthor{\bsnm{Shaikovski}, \binits{G.}},
\bauthor{\bsnm{Zelechowski}, \binits{M.}},
\bauthor{\bsnm{Severson}, \binits{K.}},
\bauthor{\bsnm{Zimmermann}, \binits{E.}},
\bauthor{\bsnm{Hall}, \binits{J.}},
\bauthor{\bsnm{Tenenholtz}, \binits{N.}},
\bauthor{\bsnm{Fusi}, \binits{N.}}, \betal:
\batitle{A foundation model for clinical-grade computational pathology and rare cancers detection}.
\bjtitle{Nature medicine}
\bvolume{30}(\bissue{10}),
\bfpage{2924}--\blpage{2935}
(\byear{2024})
\end{barticle}
\endbibitem

\bibitem[\protect\citeauthoryear{Huang et~al.}{2023}]{huang2023visual}
\begin{barticle}
\bauthor{\bsnm{Huang}, \binits{Z.}},
\bauthor{\bsnm{Bianchi}, \binits{F.}},
\bauthor{\bsnm{Yuksekgonul}, \binits{M.}},
\bauthor{\bsnm{Montine}, \binits{T.J.}},
\bauthor{\bsnm{Zou}, \binits{J.}}:
\batitle{A visual--language foundation model for pathology image analysis using medical twitter}.
\bjtitle{Nature medicine}
\bvolume{29}(\bissue{9}),
\bfpage{2307}--\blpage{2316}
(\byear{2023})
\end{barticle}
\endbibitem

\bibitem[\protect\citeauthoryear{Lu et~al.}{2024a}]{lu2024visual}
\begin{barticle}
\bauthor{\bsnm{Lu}, \binits{M.Y.}},
\bauthor{\bsnm{Chen}, \binits{B.}},
\bauthor{\bsnm{Williamson}, \binits{D.F.}},
\bauthor{\bsnm{Chen}, \binits{R.J.}},
\bauthor{\bsnm{Liang}, \binits{I.}},
\bauthor{\bsnm{Ding}, \binits{T.}},
\bauthor{\bsnm{Jaume}, \binits{G.}},
\bauthor{\bsnm{Odintsov}, \binits{I.}},
\bauthor{\bsnm{Le}, \binits{L.P.}},
\bauthor{\bsnm{Gerber}, \binits{G.}}, \betal:
\batitle{A visual-language foundation model for computational pathology}.
\bjtitle{Nature Medicine}
\bvolume{30}(\bissue{3}),
\bfpage{863}--\blpage{874}
(\byear{2024})
\end{barticle}
\endbibitem

\bibitem[\protect\citeauthoryear{Lu et~al.}{2024b}]{lu2024multimodal}
\begin{barticle}
\bauthor{\bsnm{Lu}, \binits{M.Y.}},
\bauthor{\bsnm{Chen}, \binits{B.}},
\bauthor{\bsnm{Williamson}, \binits{D.F.}},
\bauthor{\bsnm{Chen}, \binits{R.J.}},
\bauthor{\bsnm{Zhao}, \binits{M.}},
\bauthor{\bsnm{Chow}, \binits{A.K.}},
\bauthor{\bsnm{Ikemura}, \binits{K.}},
\bauthor{\bsnm{Kim}, \binits{A.}},
\bauthor{\bsnm{Pouli}, \binits{D.}},
\bauthor{\bsnm{Patel}, \binits{A.}}, \betal:
\batitle{A multimodal generative ai copilot for human pathology}.
\bjtitle{Nature}
\bvolume{634}(\bissue{8033}),
\bfpage{466}--\blpage{473}
(\byear{2024})
\end{barticle}
\endbibitem

\bibitem[\protect\citeauthoryear{Hanna et~al.}{2025}]{hanna2025ethical}
\begin{barticle}
\bauthor{\bsnm{Hanna}, \binits{M.G.}},
\bauthor{\bsnm{Pantanowitz}, \binits{L.}},
\bauthor{\bsnm{Jackson}, \binits{B.}},
\bauthor{\bsnm{Palmer}, \binits{O.}},
\bauthor{\bsnm{Visweswaran}, \binits{S.}},
\bauthor{\bsnm{Pantanowitz}, \binits{J.}},
\bauthor{\bsnm{Deebajah}, \binits{M.}},
\bauthor{\bsnm{Rashidi}, \binits{H.H.}}:
\batitle{Ethical and bias considerations in artificial intelligence/machine learning}.
\bjtitle{Modern Pathology}
\bvolume{38}(\bissue{3}),
\bfpage{100686}
(\byear{2025})
\end{barticle}
\endbibitem

\bibitem[\protect\citeauthoryear{Du et~al.}{2025}]{du2025ethics}
\begin{botherref}
\oauthor{\bsnm{Du}, \binits{R.F.}},
\oauthor{\bsnm{Carbonell}, \binits{E.L.}},
\oauthor{\bsnm{Huang}, \binits{J.}},
\oauthor{\bsnm{Liu}, \binits{S.}},
\oauthor{\bsnm{Wang}, \binits{X.}},
\oauthor{\bsnm{Shen}, \binits{D.}},
\oauthor{\bsnm{Ke}, \binits{J.}}:
Ethics of foundation models in computational pathology: Overview of contemporary issues and future implications.
IEEE Transactions on Medical Imaging
(2025)
\end{botherref}
\endbibitem

\bibitem[\protect\citeauthoryear{Seyyed-Kalantari et~al.}{2021}]{seyyed2021underdiagnosis}
\begin{barticle}
\bauthor{\bsnm{Seyyed-Kalantari}, \binits{L.}},
\bauthor{\bsnm{Zhang}, \binits{H.}},
\bauthor{\bsnm{McDermott}, \binits{M.B.}},
\bauthor{\bsnm{Chen}, \binits{I.Y.}},
\bauthor{\bsnm{Ghassemi}, \binits{M.}}:
\batitle{Underdiagnosis bias of artificial intelligence algorithms applied to chest radiographs in under-served patient populations}.
\bjtitle{Nature medicine}
\bvolume{27}(\bissue{12}),
\bfpage{2176}--\blpage{2182}
(\byear{2021})
\end{barticle}
\endbibitem

\bibitem[\protect\citeauthoryear{Vaidya et~al.}{2024}]{vaidya2024demographic}
\begin{barticle}
\bauthor{\bsnm{Vaidya}, \binits{A.}},
\bauthor{\bsnm{Chen}, \binits{R.J.}},
\bauthor{\bsnm{Williamson}, \binits{D.F.}},
\bauthor{\bsnm{Song}, \binits{A.H.}},
\bauthor{\bsnm{Jaume}, \binits{G.}},
\bauthor{\bsnm{Yang}, \binits{Y.}},
\bauthor{\bsnm{Hartvigsen}, \binits{T.}},
\bauthor{\bsnm{Dyer}, \binits{E.C.}},
\bauthor{\bsnm{Lu}, \binits{M.Y.}},
\bauthor{\bsnm{Lipkova}, \binits{J.}}, \betal:
\batitle{Demographic bias in misdiagnosis by computational pathology models}.
\bjtitle{Nature Medicine}
\bvolume{30}(\bissue{4}),
\bfpage{1174}--\blpage{1190}
(\byear{2024})
\end{barticle}
\endbibitem

\bibitem[\protect\citeauthoryear{Yang et~al.}{2025}]{yang2025demographic}
\begin{barticle}
\bauthor{\bsnm{Yang}, \binits{Y.}},
\bauthor{\bsnm{Liu}, \binits{Y.}},
\bauthor{\bsnm{Liu}, \binits{X.}},
\bauthor{\bsnm{Gulhane}, \binits{A.}},
\bauthor{\bsnm{Mastrodicasa}, \binits{D.}},
\bauthor{\bsnm{Wu}, \binits{W.}},
\bauthor{\bsnm{Wang}, \binits{E.J.}},
\bauthor{\bsnm{Sahani}, \binits{D.}},
\bauthor{\bsnm{Patel}, \binits{S.}}:
\batitle{Demographic bias of expert-level vision-language foundation models in medical imaging}.
\bjtitle{Science Advances}
\bvolume{11}(\bissue{13}),
\bfpage{0305}
(\byear{2025})
\end{barticle}
\endbibitem

\bibitem[\protect\citeauthoryear{Gustafsson and Rantalainen}{2024}]{gustafsson2024evaluating}
\begin{botherref}
\oauthor{\bsnm{Gustafsson}, \binits{F.K.}},
\oauthor{\bsnm{Rantalainen}, \binits{M.}}:
Evaluating computational pathology foundation models for prostate cancer grading under distribution shifts.
arXiv preprint arXiv:2410.06723
(2024)
\end{botherref}
\endbibitem

\bibitem[\protect\citeauthoryear{Campanella et~al.}{2025}]{campanella2025clinical}
\begin{barticle}
\bauthor{\bsnm{Campanella}, \binits{G.}},
\bauthor{\bsnm{Chen}, \binits{S.}},
\bauthor{\bsnm{Singh}, \binits{M.}},
\bauthor{\bsnm{Verma}, \binits{R.}},
\bauthor{\bsnm{Muehlstedt}, \binits{S.}},
\bauthor{\bsnm{Zeng}, \binits{J.}},
\bauthor{\bsnm{Stock}, \binits{A.}},
\bauthor{\bsnm{Croken}, \binits{M.}},
\bauthor{\bsnm{Veremis}, \binits{B.}},
\bauthor{\bsnm{Elmas}, \binits{A.}}, \betal:
\batitle{A clinical benchmark of public self-supervised pathology foundation models}.
\bjtitle{Nature Communications}
\bvolume{16}(\bissue{1}),
\bfpage{3640}
(\byear{2025})
\end{barticle}
\endbibitem

\bibitem[\protect\citeauthoryear{Ma et~al.}{2025}]{ma2025pathbench}
\begin{botherref}
\oauthor{\bsnm{Ma}, \binits{J.}},
\oauthor{\bsnm{Xu}, \binits{Y.}},
\oauthor{\bsnm{Zhou}, \binits{F.}},
\oauthor{\bsnm{Wang}, \binits{Y.}},
\oauthor{\bsnm{Jin}, \binits{C.}},
\oauthor{\bsnm{Guo}, \binits{Z.}},
\oauthor{\bsnm{Wu}, \binits{J.}},
\oauthor{\bsnm{Tang}, \binits{O.K.}},
\oauthor{\bsnm{Zhou}, \binits{H.}},
\oauthor{\bsnm{Wang}, \binits{X.}}, et al.:
Pathbench: A comprehensive comparison benchmark for pathology foundation models towards precision oncology.
arXiv preprint arXiv:2505.20202
(2025)
\end{botherref}
\endbibitem

\bibitem[\protect\citeauthoryear{Komura and Ishikawa}{2020}]{komura2020histology}
\begin{botherref}
\oauthor{\bsnm{Komura}, \binits{D.}},
\oauthor{\bsnm{Ishikawa}, \binits{S.}}:
Histology images from uniform tumor regions in tcga whole slide images.
(No Title)
(2020)
\end{botherref}
\endbibitem

\bibitem[\protect\citeauthoryear{Ilse et~al.}{2018}]{ilse2018attention}
\begin{bchapter}
\bauthor{\bsnm{Ilse}, \binits{M.}},
\bauthor{\bsnm{Tomczak}, \binits{J.}},
\bauthor{\bsnm{Welling}, \binits{M.}}:
\bctitle{Attention-based deep multiple instance learning}.
In: \bbtitle{International Conference on Machine Learning},
pp. \bfpage{2127}--\blpage{2136}
(\byear{2018}).
\bcomment{PMLR}
\end{bchapter}
\endbibitem

\bibitem[\protect\citeauthoryear{Komura et~al.}{2022}]{komura2022universal}
\begin{botherref}
\oauthor{\bsnm{Komura}, \binits{D.}},
\oauthor{\bsnm{Kawabe}, \binits{A.}},
\oauthor{\bsnm{Fukuta}, \binits{K.}},
\oauthor{\bsnm{Sano}, \binits{K.}},
\oauthor{\bsnm{Umezaki}, \binits{T.}},
\oauthor{\bsnm{Koda}, \binits{H.}},
\oauthor{\bsnm{Suzuki}, \binits{R.}},
\oauthor{\bsnm{Tominaga}, \binits{K.}},
\oauthor{\bsnm{Ochi}, \binits{M.}},
\oauthor{\bsnm{Konishi}, \binits{H.}}, et al.:
Universal encoding of pan-cancer histology by deep texture representations.
Cell Reports
\textbf{38}(9)
(2022)
\end{botherref}
\endbibitem

\bibitem[\protect\citeauthoryear{Tolkach et~al.}{2023}]{tolkach2023artificial}
\begin{barticle}
\bauthor{\bsnm{Tolkach}, \binits{Y.}},
\bauthor{\bsnm{Wolgast}, \binits{L.M.}},
\bauthor{\bsnm{Damanakis}, \binits{A.}},
\bauthor{\bsnm{Pryalukhin}, \binits{A.}},
\bauthor{\bsnm{Schallenberg}, \binits{S.}},
\bauthor{\bsnm{Hulla}, \binits{W.}},
\bauthor{\bsnm{Eich}, \binits{M.-L.}},
\bauthor{\bsnm{Schroeder}, \binits{W.}},
\bauthor{\bsnm{Mukhopadhyay}, \binits{A.}},
\bauthor{\bsnm{Fuchs}, \binits{M.}}, \betal:
\batitle{Artificial intelligence for tumour tissue detection and histological regression grading in oesophageal adenocarcinomas: a retrospective algorithm development and validation study}.
\bjtitle{The Lancet Digital Health}
\bvolume{5}(\bissue{5}),
\bfpage{265}--\blpage{275}
(\byear{2023})
\end{barticle}
\endbibitem

\bibitem[\protect\citeauthoryear{Oquab et~al.}{2023}]{oquab2023dinov2}
\begin{botherref}
\oauthor{\bsnm{Oquab}, \binits{M.}},
\oauthor{\bsnm{Darcet}, \binits{T.}},
\oauthor{\bsnm{Moutakanni}, \binits{T.}},
\oauthor{\bsnm{Vo}, \binits{H.}},
\oauthor{\bsnm{Szafraniec}, \binits{M.}},
\oauthor{\bsnm{Khalidov}, \binits{V.}},
\oauthor{\bsnm{Fernandez}, \binits{P.}},
\oauthor{\bsnm{Haziza}, \binits{D.}},
\oauthor{\bsnm{Massa}, \binits{F.}},
\oauthor{\bsnm{El-Nouby}, \binits{A.}}, et al.:
Dinov2: Learning robust visual features without supervision.
arXiv preprint arXiv:2304.07193
(2023)
\end{botherref}
\endbibitem

\bibitem[\protect\citeauthoryear{Hu et~al.}{2021}]{hu2021lora}
\begin{botherref}
\oauthor{\bsnm{Hu}, \binits{E.J.}},
\oauthor{\bsnm{Shen}, \binits{Y.}},
\oauthor{\bsnm{Wallis}, \binits{P.}},
\oauthor{\bsnm{Allen-Zhu}, \binits{Z.}},
\oauthor{\bsnm{Li}, \binits{Y.}},
\oauthor{\bsnm{Wang}, \binits{S.}},
\oauthor{\bsnm{Wang}, \binits{L.}},
\oauthor{\bsnm{Chen}, \binits{W.}}:
Lora: Low-rank adaptation of large language models.
arXiv preprint arXiv:2106.09685
(2021)
\end{botherref}
\endbibitem

\end{thebibliography}

\begin{sidewaystable}[htbp]
\renewcommand{\tablename}{Extended Table}
  \centering
  \caption{Details of all datasets used in this study, including task type, associated cancer type, source datasets, image scale, label, number of samples and classes.}
    \begin{tabular}{c p{2.2cm} p{3.5cm} p{2.5cm} p{2cm} c p{2.8cm} c c}
    \toprule
          & Name  & Task  & Cancer type & Source dataset & Tile/WSI & Label & \multicolumn{1}{p{4.19em}}{Number} & Class No \\
    \midrule
    1     & LUAD-AGE & age classification & LUAD  & TCGA-LUAD & WSI   & age group & 522 & 4 \\
    2     & KIRC-AGE & age classification & KIRC  & TCGA-KIRC & WSI   & age group & 519 & 4 \\
    3     & COAD-AGE & age classification & COAD  & TCGA-COAD & WSI   & age group & 457 & 4 \\
    \midrule
    4     & LUAD-GEN & gender classification & LUAD  & TCGA-LUAD & WSI   & gender & 541 & 2 \\
    5     & KIRC-GEN & gender classification & KIRC  & TCGA-KIRC & WSI   & gender & 519 & 2 \\
    6     & COAD-GEN & gender classification & COAD  & TCGA-COAD & WSI   & gender & 457 & 2 \\
    \midrule
    7    & LUAD-RACE & race classification & LUAD  & TCGA-LUAD & WSI   & race  & 468 & 2 \\
    8    & KIRC-RACE & race classification & KIRC  & TCGA-KIRC & WSI   & race  & 504 & 2 \\
    9    & COAD-RACE & race classification & COAD  & TCGA-COAD & WSI   & race  & 273 & 2 \\
    \midrule
    10    & LUNG-INS & institution classification & LUSC/LAUD & TCGA-UT & Tile  & institution name & 21480 & 12 \\
    11    & RCC-INS & institution classification & KIRC/KICH/KIRP & TCGA-UT & Tile  & institution name & 13110 & 8 \\
    12    & BRCA-INS & institution classification & BRCA  & TCGA-UT & Tile  & institution name & 20950 & 13 \\
    13    & COAD-INS & institution classification & COAD  & TCGA-UT & Tile  & institution name & 6620  & 9 \\
    14    & IFFM-pretrain & IFFM pretraining & pan-cancer & TCGA-UT & Tile  & institution name & 210960 & 103 \\
    15    & OA-INS & IFFM testing & sophageal adenocarcinomas & OA    & Tile  & institution name & 334533 & 3 \\
    \midrule
    16    & LUNG-tile & subtype classification & LUSC/LAUD & TCGA-UT & Tile  & cancer subtype & 22500 & 2 \\
    17    & RCC-tile & subtype classification & KIRC/KICH/KIRP & TCGA-UT & Tile  & cancer subtype & 17550 & 3 \\
    18    & LUNG-wsi & subtype classification & LUSC/LAUD & TCGA  & WSI   & cancer subtype & 1053 & 2 \\
    19    & RCC-wsi & subtype classification & KIRC/KICH/KIRP & TCGA  & WSI   & cancer subtype & 937 & 3 \\
    20    & STS-wsi & subtype classification & 6 subtypes  & private & WSI   & cancer subtype & 1072 & 2 \\
    \midrule
    21    & LUAD-TP53 & gene mutation prediction & LUAD  & TCGA-LUAD & WSI   & mutation state &  528 &2 \\
    22    & GBM\&LGG-IDH1 & gene mutation prediction & GBM/LGG & TCGA-GBM/TCGA-LGG & WSI   & mutation state & 1372 &2 \\
    \midrule
    23    & LUAD-SUR & survival analysis & LUAD  & TCGA-LUAD & WSI   & survival time \& censorship & 465 & / \\
    24    & KIRC-SUR & survival analysis & KIRC  & TCGA-KIRC & WSI   & survival time \& censorship & 503  & / \\
    25    & UCEC-SUR & survival analysis & UCEC  & TCGA-UCEC & WSI   & survival time \& censorship & 502  & / \\
    26    & TCGA-pan-cancer-RETR & image retrieval  & pan-cancer & TCGA  & Tile  & cancer type & 8736 & 32 \\
    \bottomrule
    \end{tabular}%
    \label{tab:dataset}%
\end{sidewaystable}%

\begin{table}[htbp]
\renewcommand{\tablename}{Extended Table}
  \centering
  \caption{Prediction accuracy of demographic attributes (gender, race, and age) from PFM-derived feature embeddings. Gender and race are formulated as binary classification tasks, while age is categorized into four groups.}
    \begin{tabular}{clcccccc}
    \toprule
          &       & \multicolumn{2}{c}{UNI} & \multicolumn{2}{c}{Prov-GigaPath} & \multicolumn{2}{c}{Virchow} \\
          \cmidrule(l){3-4} \cmidrule(l){5-6} \cmidrule(l){7-8}
          &       & ACC   & F1-score & ACC   & F1-score & ACC   & F1-score \\
    \midrule
    \multirow{3}[2]{*}{Gender} & LUAD-GEN & 0.6377  & 0.6314  & 0.6266  & 0.6161  & 0.6212  & 0.6129  \\
          & KIRC-GEN & 0.6859  & 0.6303  & 0.6743  & 0.6269  & 0.6319  & 0.5096  \\
          & COAD-GEN & 0.5273  & 0.4865  & 0.5580  & 0.5314  & 0.5274  & 0.4769  \\
    \midrule
    \multirow{3}[2]{*}{Race} & LUAD-RACE & 0.8867  & 0.6388  & 0.8718  & 0.5113  & 0.8696  & 0.5329  \\
          & KIRC-RACE & 0.9306  & 0.8255  & 0.9186  & 0.7816  & 0.9227  & 0.7834  \\
          & COAD-RACE & 0.8718  & 0.7938  & 0.8532  & 0.7812  & 0.8792  & 0.8581  \\
    \midrule
    \multirow{3}[2]{*}{Age} & LUAD-AGE & 0.3622  & 0.3603  & 0.3564  & 0.3456  & 0.2836  & 0.2671  \\
          & KIRC-AGE & 0.2871  & 0.2639  & 0.2871  & 0.2711  & 0.2852  & 0.2522  \\
          & COAD-AGE & 0.3020  & 0.2648  & 0.2713 & 0.2291  & 0.3040  & 0.2623  \\
    \bottomrule
    \end{tabular}%
  \label{tab:demographic_cls}%
\end{table}%

\begin{table}[h]
\renewcommand{\tablename}{Extended Table}
      \caption{The performance of institution classification when utilizing features from three PFMs. Class No. denotes the number of institutions.}
    \begin{tabular}{lccccccc}
    \toprule
          &       & \multicolumn{2}{c}{UNI} & \multicolumn{2}{c}{Prov-GigaPath} & \multicolumn{2}{c}{Virchow} \\
        \cmidrule(l){3-4} \cmidrule(l){5-6} \cmidrule(l){7-8}
          & Class No. & ACC   & F1-score & ACC   & F1-score & ACC   & F1-score \\
    \midrule
    LUNG-INS & 12    & 0.7169  & 0.4867  & 0.5911  & 0.3127  & 0.5202  & 0.2686  \\
    RCC-INS & 8     & 0.7871  & 0.6204  & 0.7683  & 0.5601  & 0.7423  & 0.5130  \\
    \hline
    BRCA-INS w/o norm & 13    & 0.6750  & 0.5430  & 0.7370  & 0.6717  & 0.5628  & 0.4128  \\
    BRCA-INS w/ norm & 13    & 0.4998  & 0.3575  & 0.5792  & 0.5026  & 0.4705  & 0.3568  \\
    COAD-INS w/o norm & 9     & 0.6600  & 0.4954  & 0.6438  & 0.4811  & 0.5368  & 0.3415  \\
    COAD-INS w/ norm & 9     & 0.5611  & 0.4085  & 0.5442  & 0.3867  & 0.5490  & 0.4001  \\
    \bottomrule
    \end{tabular}
  \label{tab:ins_cls}
\end{table}

\begin{table}[htbp]
\renewcommand{\tablename}{Extended Table}
    \caption{Patch classification performance of the LUNG-patch dataset across three OOD settings. }
    \begin{tabular}{lcccccc}
    \toprule
          & \multicolumn{2}{c}{UNI} & \multicolumn{2}{c}{Prov-GigaPath} & \multicolumn{2}{c}{Virchow} \\
          \cmidrule(l){2-3} \cmidrule(l){4-5} \cmidrule(l){6-7}
          & \multicolumn{1}{l}{ACC} & \multicolumn{1}{l}{F1-score} & \multicolumn{1}{l}{ACC} & \multicolumn{1}{l}{F1-score} & \multicolumn{1}{l}{ACC} & \multicolumn{1}{l}{F1-score} \\
    \midrule
    Baseline 1\&2 & 0.8934 & 0.8924 & 0.8958 & 0.8951 & 0.8883 & 0.8874 \\
    OOD1  & 0.8481 & 0.8470 & 0.8648 & 0.8634 & 0.8642 & 0.8625 \\
    OOD2  & 0.8095 & 0.8057 & 0.8082 & 0.8043 & 0.7612 & 0.7488 \\
    \midrule
    Baseline 3 & 0.7410 & 0.6622 & 0.9010 & 0.8997 & 0.8855 & 0.8840 \\
    OOD3  & 0.4455 & 0.3536 & 0.5809 & 0.5798 & 0.6082 & 0.6077 \\
    \bottomrule
    \end{tabular}%
  \label{tab:ood_patch_lung}%
\end{table}%

\begin{table}[htbp]
\renewcommand{\tablename}{Extended Table}
    \caption{Patch classification performance of the RCC-patch dataset across three OOD settings. }
    \begin{tabular}{lcccccc}
    \toprule
          & \multicolumn{2}{c}{UNI} & \multicolumn{2}{c}{Prov-GigaPath} & \multicolumn{2}{c}{Virchow} \\
          \cmidrule(l){2-3} \cmidrule(l){4-5} \cmidrule(l){6-7}
          & \multicolumn{1}{l}{ACC} & \multicolumn{1}{l}{F1-score} & \multicolumn{1}{l}{ACC} & \multicolumn{1}{l}{F1-score} & \multicolumn{1}{l}{ACC} & \multicolumn{1}{l}{F1-score} \\
    \midrule
    Baseline 1\&2 & 0.9394 & 0.9259 & 0.7265 & 0.4876 & 0.8419 & 0.7347 \\
    OOD1  & 0.8968 & 0.8638 & 0.7163 & 0.4513 & 0.7806 & 0.5976 \\
    OOD2  & 0.8160 & 0.7787 & 0.6897 & 0.5149 & 0.8010 & 0.7543 \\
    \midrule
    Baseline 3 & 0.9437 & 0.9306 & 0.8233 & 0.6634 & 0.7776 & 0.5736 \\
    OOD3  & 0.7219 & 0.6946 & 0.5208 & 0.4099 & 0.5256 & 0.4137 \\
    \bottomrule
    \end{tabular}%
  \label{tab:ood_patch_rcc}%
\end{table}%

\begin{table}[htbp]
\renewcommand{\tablename}{Extended Table}
    \caption{WSI classification performance across three datasets for subtype classification.}
    \begin{tabular}{clcccccc}
    \toprule
          &       & \multicolumn{2}{c}{UNI} & \multicolumn{2}{c}{Prov-GigaPath} & \multicolumn{2}{c}{Virchow} \\
          \cmidrule(l){3-4} \cmidrule(l){5-6} \cmidrule(l){7-8}
          &       & ACC   & F1-score & ACC   & F1-score & ACC   & F1-score \\
    \midrule
    \multirow{4}[2]{*}{LUNG-subtype} & Baseline 1\&2 & 0.9478  & 0.9476  & 0.9393  & 0.9391  & 0.9335  & 0.9333  \\
          & OOD1  & 0.9333  & 0.9223  & 0.9249  & 0.9127  & 0.9033  & 0.8883  \\
          & Baseline 3 & 0.9425  & 0.9416  & 0.9378  & 0.9368  & 0.9213  & 0.9195  \\
          & OOD3  & 0.7960  & 0.7943  & 0.7967  & 0.7955  & 0.7728  & 0.7624  \\
    \midrule
    \multirow{4}[2]{*}{RCC-subtype} & Baseline 1\&2 & 0.9477  & 0.9392  & 0.9520  & 0.9409  & 0.9435  & 0.9316  \\
          & OOD1  & 0.8674  & 0.8218  & 0.8663  & 0.8174  & 0.8826  & 0.8479  \\
          & Baseline 3 & 0.9472  & 0.9309  & 0.9366  & 0.9117  & 0.9525  & 0.9356  \\
          & OOD3  & 0.6919  & 0.6080  & 0.7387  & 0.6576  & 0.6661  & 0.5865  \\
    \midrule
    \multirow{4}[2]{*}{SARC-subtype} & Baseline 1\&2 & 0.9590  & 0.9572  & 0.9552  & 0.9537  & 0.9524  & 0.9503  \\
          & OOD1  & 0.8749  & 0.8618  & 0.8455  & 0.8248  & 0.8682  & 0.8540  \\
          & Baseline 3 & 0.9396  & 0.9395  & 0.9470  & 0.9469  & 0.9190  & 0.9188  \\
          & OOD3  & 0.2235  & 0.2069  & 0.3068  & 0.2742  & 0.3922  & 0.3388  \\
    \bottomrule
    \end{tabular}
  \label{tab:ood_wsi_subtype}
\end{table}

\begin{table}[htbp]
\renewcommand{\tablename}{Extended Table}
    \caption{WSI classification performance across two datasets for gene mutation prediction.}
    \begin{tabular}{clcccccc}
    \toprule
          &       & \multicolumn{2}{c}{UNI} & \multicolumn{2}{c}{Prov-GigaPath} & \multicolumn{2}{c}{Virchow} \\
          \cmidrule(l){3-4} \cmidrule(l){5-6} \cmidrule(l){7-8}
          &       & ACC   & F1-score & ACC   & F1-score & ACC   & F1-score \\
    \midrule
    \multirow{4}[2]{*}{LUAD-TP53} & Baseline 1\&2 & 0.6875  & 0.6854  & 0.6744  & 0.6732  & 0.6705  & 0.6697  \\
          & OOD1  & 0.5009  & 0.4860  & 0.5197  & 0.4870  & 0.5265  & 0.5140  \\
          & Baseline 3 & 0.6704  & 0.6679  & 0.7111  & 0.7081  & 0.6259  & 0.6232  \\
          & OOD3  & 0.2919  & 0.2889  & 0.3292  & 0.3252  & 0.2756  & 0.2742  \\
    \midrule
    \multirow{4}[2]{*}{GBM\&LGG-IDH1} & Baseline 1\&2 & 0.9177  & 0.9174  & 0.9235  & 0.9233  & 0.9016  & 0.9011  \\
          & OOD1  & 0.7594  & 0.7380  & 0.8688  & 0.8467  & 0.5938  & 0.5898  \\
          & Baseline 3 & 0.9112  & 0.9110  & 0.9197  & 0.9196  & 0.8942  & 0.8940  \\
          & OOD3  & 0.2924  & 0.2889  & 0.3527  & 0.3457  & 0.1876  & 0.1872  \\
    \bottomrule
    \end{tabular}
  \label{tab:ood_wsi_gene}
\end{table}

\begin{table}[htbp]
\renewcommand{\tablename}{Extended Table}
    \caption{WSI-based survival prediction performance across three datasets.}
    \begin{tabular}{clccc}
    \toprule
          &       & UNI   & Prov-GigaPath & Virchow \\
    \midrule
    \multirow{2}[2]{*}{LUAD-sur} & Baseline 1\&2 & 0.6059  & 0.6284  & 0.6345  \\
          & OOD1  & 0.5970  & 0.6158  & 0.5901  \\
    \midrule
    \multirow{2}[2]{*}{KIRC-sur} & Baseline 1\&2 & 0.7164  & 0.7383  & 0.7058  \\
          & OOD1  & 0.6950  & 0.7144  & 0.6908  \\
    \midrule
    \multirow{2}[2]{*}{UCEC-sur} & Baseline 1\&2 & 0.7220  & 0.7161  & 0.7252  \\
          & OOD1  & 0.6623  & 0.7025  & 0.6891  \\
    \bottomrule
    \end{tabular}%
  \label{tab:ood_wsi_sur}%
\end{table}%

\begin{table}[htbp]
\renewcommand{\tablename}{Extended Table}
    \caption{Image retrieval performance when using three PFMs.}
    \begin{tabular}{clcccc}
    \toprule
          &       & Acc@1 & Acc@3 & Acc@5 & MVAcc@5 \\
    \midrule
    \multirow{4}[2]{*}{UNI} & Baseline 1\&2 & 0.7406  & 0.8326  & 0.8686  & 0.7743  \\
          & OOD1  & 0.5690  & 0.6880  & 0.7385  & 0.6108  \\
          & Baseline 3 & 0.7365  & 0.8291  & 0.8660  & 0.7702  \\
          & OOD3  & 0.5648  & 0.6887  & 0.7434  & 0.6086  \\
    \midrule
    \multirow{4}[2]{*}{Prov-GigaPath} & Baseline 1\&2 & 0.7268  & 0.8178  & 0.8544  & 0.7586  \\
          & OOD1  & 0.5518  & 0.6686  & 0.7199  & 0.5913  \\
          & Baseline 3 & 0.7215  & 0.8133  & 0.8504  & 0.7519  \\
          & OOD3  & 0.5475  & 0.6691  & 0.7232  & 0.5866  \\
    \midrule
    \multirow{4}[2]{*}{Virchow} & Baseline 1\&2 & 0.6988  & 0.8021  & 0.8428  & 0.7344  \\
          & OOD1  & 0.5571  & 0.6789  & 0.7308  & 0.5969  \\
          & Baseline 3 & 0.6936  & 0.7976  & 0.8393  & 0.7284  \\
          & OOD3  & 0.5439  & 0.6743  & 0.7313  & 0.5896  \\
    \bottomrule
    \end{tabular}%
  \label{tab:ood_retri}%
\end{table}%

\begin{table}[htbp]
\renewcommand{\tablename}{Extended Table}
  \caption{Model performance stratified by gender on two patch-level datasets for subtype classification.}
    \begin{tabular}{clcccccc}
    \toprule
          &       & \multicolumn{2}{c}{UNI} & \multicolumn{2}{c}{Prov-GigaPath} & \multicolumn{2}{c}{Virchow} \\
          \cmidrule(l){3-4} \cmidrule(l){5-6} \cmidrule(l){7-8}
          &       & ACC   & F1-score & ACC   & F1-score & ACC   & F1-score \\
    \midrule
    \multirow{3}[2]{*}{LUNG-tile} & male & 0.8842  & 0.9019  & 0.8909  & 0.9080  & 0.8811  & 0.8994  \\
          & female & 0.9057  & 0.8250  & 0.9034  & 0.8252  & 0.8989  & 0.8175  \\
          & $|\Delta|$ & 0.0215  & 0.0769  & 0.0125  & 0.0828  & 0.0178  & 0.0819  \\
    \midrule
    \multirow{3}[2]{*}{RCC-tile} & male & 0.9361  & 0.9203  & 0.6876  & 0.4583  & 0.8161  & 0.7062  \\
          & female & 0.9465  & 0.9282  & 0.8057  & 0.5392  & 0.8938  & 0.7827  \\
          &  $|\Delta|$ & 0.0104 & 0.0079 & 0.1181 & 0.0809 & 0.0777 & 0.0765 \\
    \bottomrule
    \end{tabular}%
  \label{tab:fairness_gender_patch}%
\end{table}%

\begin{table}[htbp]
\renewcommand{\tablename}{Extended Table}
  \caption{Model performance stratified by gender on two slide-level datasets for subtype classification.}
    \begin{tabular}{clcccccc}
    \toprule
          &       & \multicolumn{2}{c}{UNI} & \multicolumn{2}{c}{Prov-GigaPath} & \multicolumn{2}{c}{Virchow} \\
          \cmidrule(l){3-4} \cmidrule(l){5-6} \cmidrule(l){7-8}
          &       & ACC   & F1-score & ACC   & F1-score & ACC   & F1-score \\
    \midrule
    \multirow{3}[2]{*}{LUNG-wsi} & male & 0.9513  & 0.9592  & 0.9328  & 0.9440  & 0.9304  & 0.9415  \\
          & female & 0.9426  & 0.9026  & 0.9475  & 0.9109  & 0.9389  & 0.8956  \\
          & $|\Delta|$ & 0.0087  & 0.0566  & 0.0147  & 0.0331  & 0.0085  & 0.0459  \\
    \midrule
    \multirow{3}[2]{*}{RCC-wsi} & male & 0.9453  & 0.9253  & 0.9533  & 0.9322  & 0.9581  & 0.9414  \\
          & female & 0.9526  & 0.9483  & 0.9491  & 0.9431  & 0.9141  & 0.8967  \\
          & $|\Delta|$ & 0.0073 & 0.023 & 0.0042 & 0.0109 & 0.044 & 0.0447 \\
    \bottomrule
    \end{tabular}%
  \label{tab:fairness_gender_wsi_subtype}%
\end{table}%

\begin{table}[htbp]
\renewcommand{\tablename}{Extended Table}
  \caption{Model performance stratified by gender on two slide-level datasets for gene mutation prediction.}
    \begin{tabular}{clcccccc}
    \toprule
          &       & \multicolumn{2}{c}{UNI} & \multicolumn{2}{c}{Prov-GigaPath} & \multicolumn{2}{c}{Virchow} \\
          \cmidrule(l){3-4} \cmidrule(l){5-6} \cmidrule(l){7-8}
          &       & ACC   & F1-score & ACC   & F1-score & ACC   & F1-score \\
    \midrule
    \multirow{3}[2]{*}{LUNG-TP53} & male & 0.6967 & 0.6857 & 0.6862 & 0.6716 & 0.7263 & 0.7178 \\
          & female & 0.6834 & 0.6679 & 0.6688 & 0.6689 & 0.6316 & 0.6286 \\
          & $|\Delta|$ & 0.0133  & 0.0178  & 0.0174  & 0.0027  & 0.0947  & 0.0892  \\
    \midrule
    \multirow{3}[2]{*}{GBM\&LGG-IDH1} & male & 0.9199  & 0.9221  & 0.9258  & 0.9251  & 0.9018  & 0.9062  \\
          & female & 0.9147  & 0.9227  & 0.9204  & 0.9271  & 0.9011  & 0.9093  \\
          & $|\Delta|$ & 0.0052 & 0.0006 & 0.0054 & 0.002 & 0.0007 & 0.0031 \\
    \bottomrule
    \end{tabular}%
  \label{tab:fairness_gender_wsi_gene}%
\end{table}%

\begin{table}[htbp]
\renewcommand{\tablename}{Extended Table}
  \caption{Model performance stratified by race on two patch-level datasets for subtype classification. * denotes black or african american.}
    \begin{tabular}{clcccccc}
    \toprule
          &       & \multicolumn{2}{c}{UNI} & \multicolumn{2}{c}{Prov-GigaPath} & \multicolumn{2}{c}{Virchow} \\
          \cmidrule(l){3-4} \cmidrule(l){5-6} \cmidrule(l){7-8}
          &       & ACC   & F1-score & ACC   & F1-score & ACC   & F1-score \\
    \midrule
    \multirow{3}[1]{*}{LUNG-tile} & white & 0.8904  & 0.8681  & 0.8930  & 0.8735  & 0.8844  & 0.8629  \\
          & black* & 0.8877  & 0.8385  & 0.8829  & 0.8416  & 0.8807  & 0.8260  \\
          & $|\Delta|$ & 0.0027  & 0.0296  & 0.0101  & 0.0319  & 0.0037  & 0.0369  \\
    \midrule
    \multirow{3}[1]{*}{RCC-tile} & white & 0.9451  & 0.9293  & 0.7637  & 0.5030  & 0.8628  & 0.7468  \\
          & black* & 0.9016  & 0.8490  & 0.4773  & 0.3837  & 0.7032  & 0.6546  \\
          & $|\Delta|$ & 0.0435 & 0.0803 & 0.2864 & 0.1193 & 0.1596 & 0.0922 \\
    \bottomrule
    \end{tabular}%
  \label{tab:fairness_race_patch}%
\end{table}%

\begin{table}[h]
\renewcommand{\tablename}{Extended Table}
  \caption{Model performance stratified by race on two slide-level datasets for subtype classification. * denotes black or african american.}
    \begin{tabular}{clcccccc}
    \toprule
          &       & \multicolumn{2}{c}{UNI} & \multicolumn{2}{c}{Prov-GigaPath} & \multicolumn{2}{c}{Virchow} \\
          \cmidrule(l){3-4} \cmidrule(l){5-6} \cmidrule(l){7-8}
          &     & ACC   & F1-score & ACC   & F1-score & ACC   & F1-score \\
    \midrule
    \multirow{3}[1]{*}{LUNG-wsi} & white & 0.9440  & 0.9434  & 0.9375  & 0.9369  & 0.9330  & 0.9323  \\
          & black* & 0.9389  & 0.9340  & 0.9390  & 0.9322  & 0.8904  & 0.8857  \\
          & $|\Delta|$ & 0.0051  & 0.0094  & 0.0015  & 0.0047  & 0.0426  & 0.0466  \\
    \midrule
    \multirow{3}[1]{*}{RCC-wsi} & white & 0.9645  & 0.9546  & 0.9633  & 0.9494  & 0.9633  & 0.9516  \\
          & black* & 0.8587  & 0.8350  & 0.8942  & 0.8934  & 0.8551  & 0.7896  \\
          & $|\Delta|$ & 0.1058  & 0.1196  & 0.0691  & 0.0560  & 0.1082  & 0.1620  \\
    \bottomrule
    \end{tabular}%
  \label{tab:fairness_race_wsi_subtype}%
\end{table}%

\begin{table}[h]
\renewcommand{\tablename}{Extended Table}
  \caption{Model performance stratified by race on two slide-level datasets for gene mutation prediction. * denotes black or african american.}
    \begin{tabular}{clcccccc}
    \toprule
          &       & \multicolumn{2}{c}{UNI} & \multicolumn{2}{c}{Prov-GigaPath} & \multicolumn{2}{c}{Virchow} \\
          \cmidrule(l){3-4} \cmidrule(l){5-6} \cmidrule(l){7-8}
          &       & ACC   & F1-score & ACC   & F1-score & ACC   & F1-score \\
    \midrule
    \multirow{3}[2]{*}{LUNG-TP53} & white & 0.7012  & 0.6681  & 0.6736  & 0.6563  & 0.6811  & 0.6583  \\
          & black* & 0.7193  & 0.7745  & 0.7638  & 0.8250  & 0.7352  & 0.7839  \\
          & $|\Delta|$ & 0.0181  & 0.1064  & 0.0902  & 0.1687  & 0.0541  & 0.1256  \\
    \midrule
    \multirow{3}[2]{*}{GBM\&LGG-IDH1} & white & 0.9148  & 0.9199  & 0.9236  & 0.9267  & 0.8983  & 0.9052  \\
          & black* & 0.8900  & 0.8161  & 0.8885  & 0.7942  & 0.8503  & 0.7787  \\
          & $|\Delta|$ & 0.0248 & 0.1038 & 0.0351 & 0.1325 & 0.048 & 0.1265 \\
    \bottomrule
    \end{tabular}%
  \label{tab:fairness_race_wsi_gene}%
\end{table}%

\begin{table}[htbp]
\renewcommand{\tablename}{Extended Table}
  \caption{Coefficient of variation in diagnostic accuracy across institutions under the OOD1 setting.}
    \begin{tabular}{lccccccc}
    \toprule
          & LUNG-tile & RCC-tile & LUNG-wsi & RCC-wsi & STS-wsi & LUAD-TP53 & GBM\&LGG-IDH1 \\
    \midrule
    UNI   & 0.0931  & 0.1061  & 0.1168  & 0.3920  & 0.1051  & 0.0392  & 0.0702  \\
    Prov-GigaPath & 0.0959  & 0.5288  & 0.1173  & 0.3459  & 0.0962  & 0.1063  & 0.0776  \\
    Virchow & 0.0959  & 0.0866  & 0.1489  & 0.3885  & 0.0952  & 0.0661  & 0.2375  \\
    \bottomrule
    \end{tabular}%
  \label{tab:fairness_ins_ood1}%
\end{table}%

\begin{table}[htbp]
\renewcommand{\tablename}{Extended Table}
  \caption{Coefficient of variation in diagnostic accuracy across institutions under the OOD3 setting.}
    \begin{tabular}{lccccccc}
    \toprule
          & LUNG-tile & RCC-tile & LUNG-wsi & RCC-wsi & STS-wsi & LUAD-TP53 & GBM\&LGG-IDH1 \\
    \midrule
    UNI   & 0.9865  & 0.3174  & 0.2135  & 0.4945  & 0.4426  & 0.8528  & 1.0304  \\
    Prov-GigaPath & 0.3173  & 0.6243  & 0.2306  & 0.4005  & 0.7043  & 0.7396  & 0.8703  \\
    Virchow & 0.3168  & 0.6797  & 0.2367  & 0.5446  & 0.6251  & 0.8661  & 1.4748  \\
    \bottomrule
    \end{tabular}%
  \label{tab:fairness_ins_ood3}%
\end{table}%

\begin{table}[htbp]
\renewcommand{\tablename}{Extended Table}
    \caption{Performance of institution classification when using pretrained institution-specific feature foundation model.}
    \begin{tabular}{lccc}
    \toprule
          & Class No. & ACC   & F1-score \\
    \midrule
    pretrain\_test & 103   & 0.7644 & 0.6422 \\
    LUNG-INS & 12    & 0.7621 & 0.6285 \\
    RCC-INS & 8     & 0.8183 & 0.6715 \\
    OA-INS & 3     & 0.9507 & 0.9258 \\
    \bottomrule
    \end{tabular}%
  \label{tab:iffm}%
\end{table}%

\begin{table}[htbp]
\renewcommand{\tablename}{Extended Table}
    \caption{Performance comparison of patch classification across OOD settings with stain normalization.}
    \begin{tabular}{cccccccc}
    \toprule
          &       & \multicolumn{2}{c}{UNI} & \multicolumn{2}{c}{Prov-GigaPath} & \multicolumn{2}{c}{Virchow} \\
          \cmidrule(l){3-4} \cmidrule(l){5-6} \cmidrule(l){7-8}
          & w/ norm & ACC   & F1-score & ACC   & F1-score & ACC   & F1-score \\
    \midrule
    \multirow{2}[2]{*}{OOD1} & $\times$   & 0.8968  & 0.8638  & 0.7163  & 0.4513  & 0.7806  & 0.5976  \\
          & $\checkmark$   & 0.8916  & 0.8507  & 0.7299  & 0.6615  & 0.8794  & 0.8352  \\
    \midrule
    \multirow{2}[2]{*}{OOD2} & $\times$   & 0.8160  & 0.7787  & 0.6897  & 0.5149  & 0.8010  & 0.7543  \\
          & $\checkmark$   & 0.8258  & 0.7833  & 0.6913  & 0.5132  & 0.8195  & 0.7699  \\
    \midrule
    \multirow{2}[2]{*}{OOD3} & $\times$   & 0.7219  & 0.6946  & 0.5208  & 0.4099  & 0.5256  & 0.4137  \\
          & $\checkmark$   & 0.5891  & 0.4956  & 0.6386  & 0.5684  & 0.5717  & 0.4796  \\
    \bottomrule
    \end{tabular}%
  \label{tab:ood_stain_norm}%
\end{table}%

\begin{table}[htbp]
\renewcommand{\tablename}{Extended Table}
    \caption{Institution classification performance when using different feature extractors.}
    \begin{tabular}{lcccc}
    \toprule
          & \multicolumn{2}{c}{LUNG-INS} & \multicolumn{2}{c}{RCC-INS} \\
          \cmidrule(l){2-3} \cmidrule(l){4-5}
          & ACC   & F1-score & ACC   & F1-score \\
    \midrule
    UNI   & 0.7169  & 0.4867  & 0.7871  & 0.6204  \\
    Prov-GigaPath & 0.5911  & 0.3127  & 0.7683  & 0.5601  \\
    Virchow & 0.5202  & 0.2686  & 0.7423  & 0.5130  \\
    \midrule
    CONCH-V & 0.6296  & 0.4110  & 0.6486  & 0.5014  \\
    \midrule
    ViT-S-SSL & 0.7927  & 0.7248  & 0.8797  & 0.8214  \\
    ViT-S-FSL & 0.6835  & 0.5866  & 0.8093  & 0.7258  \\
    \bottomrule
    \end{tabular}%
  \label{tab:ins_cls_ssl}%
\end{table}%

\begin{table}[htbp]
\renewcommand{\tablename}{Extended Table}
  \caption{WSI classification performance under OOD settings between CONCH-V and other PFMs on the LUNG-wsi dataset.}
    \begin{tabular}{lcccccccc}
    \toprule
          & \multicolumn{2}{c}{Baseline 1\&2} & \multicolumn{2}{c}{OOD1} & \multicolumn{2}{c}{Baseline 3} & \multicolumn{2}{c}{OOD3} \\
          \cmidrule(l){2-3} \cmidrule(l){4-5} \cmidrule(l){6-7} \cmidrule(l){8-9}
          & ACC   & F1-score & ACC   & F1-score & ACC   & F1-score & ACC   & F1-score \\
    \midrule
    UNI   & 0.9478  & 0.9476  & 0.9333  & 0.9223  & 0.9425  & 0.9416  & 0.7960  & 0.7943  \\
    Prov-Gigapath & 0.9393  & 0.9391  & 0.9249  & 0.9127  & 0.9378  & 0.9368  & 0.7967  & 0.7955  \\
    Virchow & 0.9335  & 0.9333  & 0.9033  & 0.8883  & 0.9213  & 0.9195  & 0.7728  & 0.7624  \\
    CONCH-V & 0.9297  & 0.9296  & 0.9249  & 0.9119  & 0.9330  & 0.9318  & 0.9020  & 0.9008  \\
    \bottomrule
    \end{tabular}%
  \label{tab:conch_lung}%
\end{table}%

\begin{table}[htbp]
\renewcommand{\tablename}{Extended Table}
  \caption{WSI classification performance under OOD settings between CONCH-V and other PFMs on the RCC-wsi dataset.}
    \begin{tabular}{lcccccccc}
    \toprule
          & \multicolumn{2}{c}{Baseline 1\&2} & \multicolumn{2}{c}{OOD1} & \multicolumn{2}{c}{Baseline 3} & \multicolumn{2}{c}{OOD3} \\
          \cmidrule(l){2-3} \cmidrule(l){4-5} \cmidrule(l){6-7} \cmidrule(l){8-9}
          & ACC   & F1-score & ACC   & F1-score & ACC   & F1-score & ACC   & F1-score \\
    \midrule
    UNI   & 0.9477  & 0.9392  & 0.8674  & 0.8218  & 0.9472  & 0.9309  & 0.6919  & 0.6080  \\
    Prov-Gigapath & 0.9520  & 0.9409  & 0.8663  & 0.8174  & 0.9366  & 0.9117  & 0.7387  & 0.6576  \\
    Virchow & 0.9435  & 0.9316  & 0.8826  & 0.8479  & 0.9525  & 0.9356  & 0.6661  & 0.5865  \\
    CONCH-V & 0.9445  & 0.9309  & 0.8761  & 0.8291  & 0.9314  & 0.9014  & 0.8435  & 0.7706  \\
    \bottomrule
    \end{tabular}%
  \label{tab:conch_rcc}%
\end{table}%

\begin{table}[htbp]
\renewcommand{\tablename}{Extended Table}
    \caption{Performance comparison of patch classification across OOD settings with fine-tuned PFMs.}
    \begin{tabular}{llcccccc}
    \toprule
          &       & \multicolumn{2}{c}{OOD1} & \multicolumn{2}{c}{OOD2} & \multicolumn{2}{c}{OOD3} \\
          \cmidrule(l){3-4} \cmidrule(l){5-6} \cmidrule(l){7-8} 
          &       & UNI   & UNI+LoRA & UNI   & UNI+LoRA & UNI   & UNI+LoRA \\
    \midrule
    \multirow{2}[2]{*}{LUNG-patch} & ACC   & 0.8481  & 0.8632  & 0.8095  & 0.8255  & 0.4455  & 0.6549  \\
          & F1-score & 0.8470  & 0.8619  & 0.8057  & 0.8226  & 0.3536  & 0.6545  \\
    \midrule
    \multirow{2}[2]{*}{RCC-patch} & ACC   & 0.8968  & 0.9036  & 0.8160  & 0.8757  & 0.7219  & 0.8454  \\
          & F1-score & 0.8638  & 0.8683  & 0.7787  & 0.8363  & 0.6946  & 0.8370  \\
    \bottomrule
    \end{tabular}%
  \label{tab:lora}%
\end{table}%

\end{document}